\documentclass[twoside,11pt]{article}

\usepackage{blindtext}

\usepackage{jmlr2e}

\usepackage{lastpage}
\jmlrheading{xx}{xx}{1-\pageref{LastPage}}{x/xx; Revised x/xx}{x/xx}{xx-xxxx}{$^*.$~Corresponding Author}

\ShortHeadings{Distributed Linear  Regression with Compositional Covariates}{Chao et al.}
\firstpageno{1}

\usepackage{mathrsfs}
\usepackage{bbm}
\usepackage{amsmath}
\usepackage{amssymb}
\usepackage{algorithm}
\usepackage{algorithmic}
\usepackage{pdflscape}
\usepackage{multirow}
\usepackage{makecell}

\newtheorem{lem}{Lemma}

\newtheorem{thm}{Theorem}[section]
\def \cude {\boldsymbol}
\def\uset{\underset}
\def\oper{\operatorname}

\begin{document}

\title{Distributed Linear  Regression with Compositional Covariates}

\author{\name Yue Chao \email 
ychaoasy@stu.suda.edu.cn\\
       \addr Department of Statistics\\
       School of Mathematical Sciences \\
       Soochow University, Suzhou, China
       \AND
       \name Lei Huang \email stahl@swjtu.edu.cn \\
       \addr Department of Statistics\\
       School of Mathematics\\
       Southwest Jiaotong University, Chengdu, China
       \AND
       \name Xuejun Ma$^*$ \email 
xuejunma@suda.edu.cn\\
       \addr Department of Statistics\\
       School of Mathematical Sciences \\
       Soochow University, Suzhou, China}

\editor{My editor}

\maketitle

\begin{abstract}
With the availability of extraordinarily huge data sets, solving the problems of distributed statistical methodology and computing
for such data sets has become increasingly crucial in the big data area. In this paper, we focus on the distributed sparse penalized
linear log-contrast model in massive compositional data. In particular, two distributed optimization techniques under centralized
and decentralized topologies are proposed for solving the two different constrained convex optimization problems. Both two proposed algorithms are based on the frameworks of Alternating Direction Method of Multipliers (ADMM)
and Coordinate Descent Method of Multipliers(CDMM, Lin et al., \textit{2014, Biometrika}). It is worth emphasizing that, in the decentralized topology, we introduce a distributed coordinate-wise descent algorithm based on Group ADMM(GADMM, Elgabli et al., \textit{2020, Journal of Machine Learning Research}) for obtaining a communication-efficient regularized estimation. Correspondingly, the convergence theories of the proposed algorithms are rigorously established under some regularity conditions. Numerical experiments on both synthetic and real data are conducted to evaluate our proposed algorithms.
\end{abstract}


\section{Introduction}
When one response variable is to be predicted by the proportions or fractions of a composition, regression models with compositional covariates are often used. A desirable regression association in terms of a log-contrast of compositional measurements was proposed by \citet{aitchison1984log}. Suppose that the full data set $\mathscr{F}_n=\{(y_i, \mathbbm{x}_i): i=1,\ldots,n\}$ consists of $n$ independent and identically distributed (i.i.d.) observations.   Given a response $\boldsymbol{Y}=(y_1,\ldots, y_n)^{\top}\in \mathbbm{R}^n$, an $n\times p$ covariate matrix $\cude{X}= (X_1,\ldots, X_p)\in \mathbbm{R}^{n\times p}$ with the unit-sum constraint consists of $n$ observations of the composition of a
mixture with $p$ components. That is, each covariate vector $\{\mathbbm{x}_i=(x_{i1},\ldots,x_{ip})^{\top}: i=1,\ldots,n\}$ of $\cude{X}$ lies in the $(p-1)-$dimensional positive Simplex $$S^{p-1}=\left\{(x_{i1},\ldots,x_{ip})\mid x_{ij>0},\sum_{j=1}^px_{ij}=1, j=1,\ldots,p \right\}.$$ The log-contrast or additive log-ratio transformation introduced by \citet{aitchison1982statistical} is adopted to $\cude{X}$, i.e.
$\cude{Z}_{-p}=(Z_1,\ldots, Z_{p-1})=\left(\log\frac{X_1}{X_p},\ldots, \log\frac{X_{p-1}}{X_p}\right),$
where $\cude{Z}_{-p}$ is the $n\times(p-1)$ log-ratio matrix.  Correspondingly, the linear log-contrast model is  formulated as follows
\begin{equation}\label{e1}
\cude{Y}= \cude{Z}_{-p}\cude{\beta}_{-p}+\cude{\epsilon},
\end{equation}
where $\cude{\beta}_{-p}=(\beta_1,\ldots, \beta_{p-1})^\top$ is the regression coefficient that need to be estimated and $\cude{\epsilon}=(\epsilon_1,\ldots, \epsilon_n)^\top$ is the independent random error vector. Notice that it is crucial to investigate the variable selection and estimation procedures for \eqref{e1}. To select the important variables from all components including reference component more conveniently, \citet{lin2014variable} re-expressed the model \eqref{e1} by setting $\beta_p=-\sum_{j=1}^{p-1}\beta_j$ as a zero-sum
constraint, that is
\begin{equation}\label{e2}
\cude{Y}=\cude{Z}\cude{\beta} + \cude{\epsilon},~\text{subject}~\text{to}~\sum_{j=1}^p\beta_j=0,
\end{equation}
where $\cude{Z}=(Z_1,\ldots, Z_p)=(\log x_{ij})\in \mathbbm{R}^{n\times p}$, $\cude{\beta}=(\beta_1,\ldots, \beta_p)^\top$.  Specifically, \citet{lin2014variable} solve the following penalized optimization problem by using Coordinate Descent Method of Multipliers(CDMM)
\begin{equation}
\cude{\widehat{\beta}}=\arg\min\limits_{\cude{\beta}}\left(\frac{1}{2n}\|\cude{Y}-\cude{Z}\cude{\beta}\|^2
+\lambda\|\cude{\beta}\|_1\right),~\text{subject}~\text{to}~\sum_{j=1}^p\beta_j=0,
\end{equation}
where $\lambda$ is a regularization parameter.\par
There exist many statistical achievements for further solving the regularization problems with compositional or subcompositional data in analysis of microbiome data(\citet{shi2016regression,wang2017structured,cao2018large,lu2019generalized,mishra2022robust,han2022robust}). For example, \cite{shi2016regression} improved the linear regression model by including several linear constraints to achieve subcompositional coherence and developed a penalized estimation methodology for selecting relevant variables and estimating the regression coefficients. \cite{wang2017structured} proposed the Tree-guided Automatic Subcomposition Selection Operator(TASSO) for conducting the tree-structured subcomposition selection and parameter estimation procedures. To tackle the issue of covariance estimation for high-dimensional compositional data, a composition-adjusted thresholding(COAT) method has been investigated by \cite{cao2018large}. Without requiring the restrictive Gaussian or sub-Gaussian assumption, \cite{li2022robust} further suggested the robust composition adjusted thresholding covariance procedure based on Huber-type M-estimation(M-COAT) to estimate the sparse covariance structure of high-dimensional compositional data. For the generalized linear models with high-dimensional compositional variables, \cite{lu2019generalized} developed a  generalized accelerated proximal gradient
algorithm to compute the estimates of regression coefficients and the statistical inference have been taken into consideration. Under the regression frameworks when containing both compositional and non-compositional covariates, \cite{mishra2022robust} proposed the Robust log-contrast Regression estimators with
Compositional Covariates (RobRegCC) procedure for outlier detection and variable selection. When the error distribution was asymmetric and heavy-tailed,  \cite{han2022robust} provided a robust signal recovery strategy for high-dimensional linear log-contrast models. For more articles on compositional data analysis, we refer to a review paper by \cite{alenazi2021review}.\par
Nevertheless, when compositional data sets have extremely large sample size, the aforementioned techniques are unable to handle such massive data sets in a single standalone machine due to the heavy burden on storage and limited computing resources. In recent years, many contributions for dealing with non-compositional massive data sets have been arisen by the researchers in the context of statistical computing and inference. A popular way is that such data sets are usually divided across multiple connected machines and calculated by many communication-efficient algorithms corresponding to different statistical problems(\cite{tan2022communication,fan2021communication,zhang2013communication,huang2021communication,liu2022communication,zhou2021communication,li2021communication,shi2021communication,jordan2018communication,yang2022communication,ma2022statistical}, etc). Among them, the distributed communication-efficient surrogate likelihood (CSL) framework proposed by \cite{jordan2018communication} aims for estimation and inference in low-dimensional regression, high-dimensional variable selection and Bayesian  statistics. It has been widely extended to various aspects including linear regression(\cite{liu2022communication,fan2021communication}), quantile regression(\cite{hu2021distributed,wang2022communication,tan2022communication},etc.),  composite quantile regression(\cite{yang2022communication,wang2021robust}), modal regression(\cite{wang2021robust}), expectile regression(\cite{pan2021distributed}), Huber regression(\cite{pan2022efficient}),  neural network(\cite{wang2022efficient}), etc. In high-dimensional settings, the estimation and variable selection procedures rather than CSL frameworks have simultaneously considered under the distributed system, refer to \cite{chen2020distributed,zhou2021communication,gu2020sparse,fan2021distributed,yu2017parallel,
volgushev2019distributed,zhu2021least,yu2022distributed}, etc. The divide-and-conquer methods for solving distributed system problems are summarized in a review paper provided by \cite{gao2022review}. It is worth pointing out that the optimization algorithms in the above-mentioned literature for the distributed data are mainly based on the frameworks of proximal Alternating Direction Method of Multipliers(ADMM, \cite{boyd2011distributed}). In addition, they solve the problems in a centralized regime. In other words, there exists a master machine, which has a link to each local machine.  To obtain a global optimal solution, the master machine collects the current local estimators calculated on the node machines after local parameter updating iteratively. Then, the aggregating parameter estimators are transmitted to the local machine for the next iteration. \par
However, such master-worker distributed framework has several drawbacks. First, the master machine is vulnerable to attackers. As a result, all informative updates from local machines that sends to the master machine can be easily obtained by attackers, which makes privacy unprotected(\cite{wu2022network}). Second, for robust operation, this centralized topology is quite fragile. In other words, if the master machine fails to work then the entire distributed system stops working. Third, due to the limited communication resources for the hardware devices in practice, the master node may require high quality for the communication network as the number of local machine grows large, which leads to the high communication cost among master machine and local workers. In addition, the issues that a large number of local machines are linked to a central machine and transmit the updated models to the central machine are unrealistic due to the limited computing power and high economic cost. More potential challenges for the master-worker distributed framework also  can be found in \cite{nedic2018network,wu2022network,elgabli2020gadmm}, etc.\par
To overcome these drawbacks, a number of researchers develop the efficient distributed algorithms in the decentralized topology(\cite{nedic2018network,elgabli2020gadmm,issaid2020communication,
 atallah2022codgrad,elgabli2020q,wu2022network}, etc). Especially, Group Alternating Direction Method of Multipliers(GADMM, \citet{elgabli2020gadmm}) is an extension to the framework of ADMM. Suppose that there exist $K$ machines. As introduced in \citet{elgabli2020gadmm}, GADMM algarithm aims to solve the optimization problem $\frac{1}{K}\sum_{k=1}^Kf_k(\cude{\alpha}_k)$ with constraints $\cude{\alpha}_k=\cude{\alpha}_{k+1},~k=1,\ldots, K-1$, where $f_{k}(\cdot)$ is the local convex, proper, and closed function. The basic principle is that the machines are split into two groups, head and tail, with each machine in the head(tail) group communicating solely with its two neighboring machines in the tail(head) group.\par
The previous works mainly focus on the distributed optimization problems with only one type of constraints. Moreover, the original GADMM algorithm as a model-free approach can only solve the parameter estimation problem without any regularization consideration for variable selection problem. In this paper, we consider the distributed estimation
  and variable selection procedures for model \eqref{e2} including non-compositional covariates and two types of constraints under the centralized and decentralized manners, which fills the gap in the fields of massive compositional data analysis. Correspondingly, we develop a efficient distributed optimization algorithms in the frameworks of ADMM and CDMM to solve the penalized  linear log-contrast
  regression models with the LASSO(\cite{tibshirani1996regression}), adaptive LASSO(\cite{zou2006adaptive}), and SCAD(\cite{fan2001variable}) penalties. More specifically,  inspired by the idea of GADMM algorithm, we introduce a distributed sparse group coordinate descent method of multipliers(DSGCDMM) for solving the two constrained optimization problem in a decentralized framework. Meanwhile, the convergence theories are established for the proposed methodologies under some assumptions.\par
  The remainder of the paper is organized as follows. Section \ref{CA} introduces a general framework for distributed penalized linear regression with compositional covariates and presents the related algorithm in a centralized regime. Section \ref{DA} presents a decentralized algorithm for the variable selection procedure in a distributed framework. The regularization parameter selection are presented in Section \ref{RPS}. In Section \ref{CT}, we present the convergence theories. Simulation experiments and real data analysis in Section \ref{NET} are conducted to demonstrate the numerical and statistical efficiency of the proposed two variable selection procedures. Finally, we conclude this article with some discussions in Section \ref{CON}. All technical details and additional numerical results are delegated to the Appendix.
    \section{A General Framework}\label{CA}
We begin by setting up our distributed regime for penalized large-scale linear regression with compositional covariates, after which
we turn to a detailed description of the distributed variable selection procedure in a master-worker system.
\subsection{Distributed Penalized Linear Regression with Compositional Covariates}
We now investigate the distributed statistical scheme for quantifying the relationship between response and compositional covariates under a master-worker framework. Without losing generality and simplicity, we assume that the response and explanatory variables are centered. In practice, compositional covariates together with $q$ additional non-compositional covariates $\cude{V}=(V_1,\ldots,V_q)\in \mathbbm{R}^{n\times q}$ are often appeared in regression analysis. Following the argument of \cite{mishra2022robust}, we consider a extended linear log-contrast model based on \eqref{e2} as presented below
\begin{equation}\label{cn}
\begin{split}
\cude{Y}&=\cude{Z}\cude{\beta} + \cude{V}\cude{\theta}+\cude{\epsilon}\\
&\equiv \cude{\Pi}\cude{\zeta} + \cude{\epsilon},~\text{subject}~\text{to}~\cude{C}^\top\cude{\zeta}=0,
\end{split}
\end{equation}
where $\cude{\Pi}=(\cude{Z}, \cude{V})$, $\cude{\zeta}=(\cude{\beta}^\top,\cude{\theta}^\top)^\top$, $\cude{C}=(\cude{1}_p^\top, \cude{0}_q^\top)^\top\in\mathbbm{R}^{(p+q)}$ with $\cude{1}_p=(1,\ldots,1)^\top\in\mathbbm{R}^p$ and $\cude{0}_q=(0,\ldots,0)^\top\in\mathbbm{R}^q$. \par
  Let $d=p+q$. For a distributed system, the full data set $\mathscr{F}_n$ are partitioned across $K$ servers(machines) $\{\mathscr{M}_k\}_{k=1}^K$. Then,
$$
\cude{Y}=(\cude{y}^\top_1,\ldots,\cude{y}^\top_K)^\top,$$
$$~\text{and}~\text{correspondingly}\\
~\cude{\Pi}=(\cude{\Pi}_1^\top,\ldots,\cude{\Pi}_K^\top)^\top,
$$
where $\cude{y}_k\in \mathbbm{R}^{n_k}$, $\cude{\Pi}_k=(\cude{\pi}_{k,1},\ldots,\cude{\pi}_{k,d})\in \mathbbm{R}^{n_k\times d}$, and $\sum_{k=1}^Kn_k=n$. In such settings, we consider following constrained convex optimization problem in a master-worker framework by applying weighted $L_1-$penalized regularization approach to model \eqref{cn},
\begin{equation}\label{optpA}
\begin{split}
\widetilde{\cude{\zeta}}&=\uset{\cude{\zeta},\{\cude{\zeta}_k\}^K_{k=1}}{\oper{argmin}}
\sum_{k=1}^{K}\frac{1}{2n_k}\|\cude{y}_k-\cude{\Pi}_k\cude{\zeta}_k\|^2
+\lambda\|\cude{\omega} \circ\cude{\zeta}\|_1,\\
&\text{subject}~\text{to}~
\begin{cases}
\cude{C}^\top\cude{\zeta}=0,\\
\cude{\zeta}_k=\cude{\zeta},
\end{cases}
k=1,\ldots, K,
\end{split}
\end{equation}
where $\cude{\zeta}$ is the global model parameter. Here, for $\mathscr{M}_k$, $\cude{\zeta}_k=(\zeta_{k,1},\ldots,\zeta_{k,d})^\top$ is the coefficient vector with respect to split data $\{\cude{y}_k, \cude{\Pi}_k\}$, $\lambda$ is a regularization parameter that dictates the size of model \eqref{cn}, the vector $\cude{\omega}=(\omega_{1},\ldots,\omega_{d})^\top$ consists of pre-specified nonnegative weights, and $\|\cude{\omega}\circ\cude{\zeta}\|_1=\sum_{j=1}^d\omega_{j}|\zeta_{j}|$ with $\circ$ standing for the Hadamard product.  It is worth mentioning that $\|\cude{\omega} \circ\cude{\zeta}\|_1$ in formulation \eqref{optpA} is a general form for the $L_1-$penalized regression with constraints. Instead, for the LASSO penalty, $\cude{\omega}$ can be specified as $\cude{\omega}=\cude{1}_d$, where $\cude{1}_d=(1,\ldots,1)^\top\in \mathbbm{R}^d$. While for the adaptive LASSO penalty, $\cude{\omega}=\left((|\widetilde{\zeta}_{j}^{\text{LASSO}}|+\frac{1}{n})^{-1}, j=1\ldots, d\right)^\top$ as a generalization of the LASSO penalty is commonly used, where $\widetilde{\cude{\zeta}}^{\text{LASSO}}=\left(\widetilde{\zeta}_{j}^{\text{LASSO}}, j=1,\ldots,d\right)^\top$ is the LASSO estimator in \cite{lin2014variable}.   If we consider the nonconvex
penalty, then SACD weight $\cude{\omega}=\left(\lambda^{-1} p_\lambda^\prime(|\overline{\zeta}_{j}|), j=1,\ldots,d\right)^\top$ is usually employed(\cite{zou2008one,liu2022communication}), where $p_\lambda^\prime(|u|)=\lambda I(|u|\leq\lambda)+\frac{(a\lambda-|u|)_+}{(a-1)I(|u|>\lambda)}$ for some $a>2$. As suggested in \cite{fan2001variable}, often $a = 3.7$ is a considerable choice. According to \cite{zou2008one}, local linear approximation(LLA) algorithm can effectively solve the SCAD-penalized regression. In this work, $\overline{\cude{\zeta}}=\left(\overline{\zeta}_{j}, j=1,\ldots, d\right)^\top$ is replaced by the current $m_{th}$ iteration $\cude{\zeta}^{m}$ in the local machines, and LLA algorithm repeatedly calculate the following constrained nonconvex optimization
\begin{equation}\label{optSCAD}
\begin{split}
\cude{\zeta}^{m+1}&=\uset{\cude{\zeta},\{\cude{\zeta}_k\}^K_{k=1}}{\oper{argmin}}
\sum_{k=1}^{K}
\frac{1}{2n_k}\|\cude{y}_k-\cude{\Pi}_k\cude{\zeta}_k\|^2
+\sum_{j=1}^dp_\lambda^\prime(|{\zeta}^{m}_{j}|)|\zeta_{j}|,\\
&\text{subject}~\text{to}~
\begin{cases}
\cude{C}^\top\cude{\zeta}=0,\\
\cude{\zeta}_k=\cude{\zeta},
\end{cases}
k=1,\ldots, K.
\end{split}
\end{equation}\par
In the next subsection, we will introduce a distributed algorithm in a centralized framework for solving problem \eqref{optpA}. The problem \eqref{optSCAD} can be also solved by combining the LLA algorithm and coordinate descent(CD) algorithm(\cite{breheny2011coordinate}) together with the distributed algorithm.
\subsection{A Distributed Sparse Coordinate Descent Method of Multipliers}\label{sec3}
We now introduce a centralized algorithm, i.e. distributed sparse coordinate descent method of multipliers(DSCDMM), to solve the weighted $L_1-$ penalized linear regression with compositional covariates in big data. Notice that there exist two kinds of constraints in the optimization problem \eqref{optpA}. To address this kind of convex optimization problem with several convex constraints, \cite{giesen2019combining} proposed a extended approach for ADMM(\cite{boyd2011distributed}), i.e. combine the superiority of ADMM to calculate the convex optimization problems in a distributed regime with the advantage of the augmented Lagrangian method to overcome constrained optimization problems. Inspired by \cite{giesen2019combining}, pick penalty parameter $\rho>0$ and denote $\cude{C}=(c_1,\ldots,c_d)^\top$, we consider the augmented Lagrangian for the optimization problem in \eqref{optpA} as
\begin{equation}
\begin{split}
\mathscr{L}_{\rho}\left(\{\cude{\zeta}_k\}^K_{k=1},\mu,\{\cude{\gamma}_k\}_{k=1}^K\right)=&\sum_{k=1}^K\frac{1}{2n_k}\|\cude{y}_k-\cude{\Pi}_k\cude{\zeta}_k\|^2
+\lambda\|\cude{\omega}\circ\cude{\zeta}\|_1+\sum_{j=1}^d\mu c_j\zeta_{j}\\
&+\frac{\rho}{2}\left(\sum_{j=1}^dc_j\zeta_{j}\right)^2+\sum_{k=1}^K\left\langle \cude{\gamma}_k, \cude{\zeta}_k-\cude{\zeta}\right\rangle
+\frac{\rho}{2}\sum_{k=1}^K\|\cude{\zeta}_k-\cude{\zeta}\|^2,
\end{split}
\end{equation}
where $\mu\in \mathbbm{R}^1$ and $\cude{\gamma}_k\in \mathbbm{R}^d$ present the Lagrangian multipliers or dual variables, $\left\langle\cdot,\cdot\right\rangle$ and $ \|\cdot\|$ stand for the inner product and $L_2-$norm in Euclidean space, respectively. Following \cite{lin2014variable} and \cite{elgabli2020gadmm}, for each machine $\mathscr{M}_k$, the updates of the primal and dual variables under the  method of multipliers are given by
\begin{equation}\label{upA}
\begin{split}
\cude{\zeta}_k^{l+1}:=\uset{\cude{\zeta}_k}{\oper{argmin}}&\frac{1}{2n_k}\|\cude{y}_k-\cude{\Pi}_k\cude{\zeta}_k\|^2+\left\langle\cude{\gamma}_k^l,\cude{\zeta}_k-\cude{\zeta}^{l}\right\rangle
+\frac{\rho}{2}\|\cude{\zeta}_k-\cude{\zeta}^{l}\|^2,
\end{split}
\end{equation}
\begin{equation}\label{upz}
\cude{\zeta}^{l+1}:=\uset{\cude{\zeta}}{\oper{argmin}}~\lambda\|\cude{\omega}\circ\cude{\zeta}\|_1+\sum_{k=1}^K\left\langle\cude{\gamma}_k^l,\cude{\zeta}_k^{l+1}-\cude{\zeta}\right\rangle
+\frac{\rho}{2}\sum_{k=1}^K\|\cude{\zeta}_k^{l+1}-\cude{\zeta}\|^2+\sum_{j=1}^d\mu c_j\zeta_{j}
+\frac{\rho}{2}\left(\sum_{j=1}^dc_j\zeta_{j}\right)^2,
\end{equation}
\begin{equation}\label{upmu}
\mu^{l+1}\leftarrow \mu^l+\rho\sum_{j=1}^dc_{j}\zeta_{j}^{l+1},
\end{equation}
\begin{equation}\label{upga}
\cude{\gamma}_{k}^{l+1}\leftarrow \cude{\gamma}_{k}^{l} + \rho(\cude{\zeta}_k^{l+1}-\cude{\zeta}^{l+1}).
\end{equation}\par
The update of each $\cude{\zeta}_k^{l+1}$ can be rewritten as the form of a Tikhonov-regularized least squares(i.e., ridge regression):
$$
\cude{\zeta}_k^{l+1}:=\uset{\cude{\zeta}_k}{\oper{argmin}}\frac{1}{2n_k}\|\cude{y}_k-\cude{\Pi}_k\cude{\zeta}_k\|^2+\frac{\rho}{2}\|\cude{\zeta}_k-\cude{\zeta}^l+\frac{1}{\rho}\cude{\gamma}^l_k\|^2,
$$
which has the analytical solution:
\begin{equation}\label{anac}
\cude{\zeta}_k^{l+1}\leftarrow (\frac{1}{n_k}\cude{\Pi}_k^\top\cude{\Pi}_k+\rho\cude{I})^{-1}(\frac{1}{n_k}\cude{\Pi}_k^\top\cude{y}_k+\rho\cude{\zeta}^l-\cude{\gamma}^l_k).
\end{equation}
  In general, updating $\cude{\zeta}$ in the subproblem \eqref{upz} is challenging due to the nondifferentiable $L_1$ terms and non-closed form solution under the zero-sum constraint. To solve this type of problems,  a class of efficient and fast algorithms named coordinate-wise descent algorithms are proposed by \cite{friedman2007pathwise}. Furthermore, \cite{lin2014variable} and \cite{gu2018admm} combine the coordinate
descent algorithm with the method of multipliers or the augmented Lagrangian method to solve the LASSO-type high-dimensional linear regression with compositional covariates and quantile regression. Hence, we prefer to apply the coordinate descent algorithm to the $\cude{\zeta}$ updates in the distributed ADMM algorithm. We coordinate-wisely solve the subproblem \eqref{upz} via updating following iterations
\begin{equation}\label{scd1}
\begin{split}
&\cude{\zeta}^{l, b+1}\leftarrow\\
& \left[\frac{\text{Shrink}\left\{
\frac{1}{\rho}\sum_{k=1}^K\gamma_{k,j}^l+\sum_{k=1}^K\zeta_{k,j}^{l+1}-\frac{1}{\rho}\mu^lc_j-c_j\sum_{m\neq j}c_m\zeta_m^{l,b + I(m<j)}
,\frac{\lambda\omega_{j}}{\rho}
\right\}}{(c_j^2+K)}\right]_{1\leq j\leq d},
\end{split}
\end{equation}
where $\text{Shrink}[\mu,\vartheta]=\text{sgn}(\mu)(|\mu|-\vartheta)_+$ is the soft thresholding operator.  We subsequently propose the \textbf{d}istributed \textbf{s}parse \textbf{c}oordinate \textbf{d}escent \textbf{m}ethod of \textbf{m}ultipliers(DSCDMM) for solving large-scale penalized linear regression with compositional covariates, which is summarized in Algorithm \ref{alg1}.\par
\begin{algorithm}[ht!]
 \renewcommand{\algorithmicrequire}{\textbf{Input:}}
 \renewcommand{\algorithmicensure}{\textbf{Output:}}
\caption{DSCDMM: Distributed sparse coordinate descent method of multipliers for solving large-scale penalized linear regression with compositional covariates.}
\label{alg1}
\begin{algorithmic}[1]
\REQUIRE Data $\{\cude{y}_{k}, \cude{\Pi}_{k}\}_{k=1}^K$ on machines $\{\mathscr{M}_k\}_{k=1}^K$, the value of constant $\rho$,  the number of rounds of communication $L$.
\ENSURE $\cude{\zeta}^L$.
\STATE {\textbf{Initialization}:
\begin{itemize}
\item Set $\mathscr{M}_1$ as the master machine.
 \item $\cude{\zeta}_k^{0}=\cude{\gamma}_k^{0}$, $\mu_k^{0}=0$ for all $k$.
 \item Calculate $\cude{\omega}$ in the first machine.
 \end{itemize}
 }
\FOR{$l=0,1,2,\ldots,L-1$}
\STATE  For each local machine, update \eqref{anac} in parallel.
\STATE Carry out the coordinate descent Steps in master machine until the convergence criterion is met.
\STATE Initialize $\cude{\zeta}^{l+1,0}=\cude{\zeta}^l$.
\FOR{$b=0,1,2,\ldots$}
\FOR{$j=1,\ldots,d$}
\STATE  Update $\zeta_j^{l, b+1}$ via \eqref{scd1}.
\ENDFOR
\STATE Set $\cude{\zeta}^{l+1}=\cude{\zeta}^{l,B}$.
\ENDFOR
\STATE The master machine updates the dual variable $\mu^{l+1}$ via \eqref{upmu}.
Broadcast $\cude{\zeta}^{l+1}$ to other local machines.
\STATE Every machine locally updates the dual variable $\cude{\gamma}_{k}^{l+1}$ \eqref{upga}.
\ENDFOR
\end{algorithmic}
\end{algorithm}\par
\section{A Decentralized Algorithm}\label{DA}
It should be noted that algorithm \ref{alg1} aims to solve the optimization problem in a centralized manner. That is, in each iteration, DSCDMM requires a master machine being connected to every node machine, then node machines calculate the local updates in parallel and send the updated variables to the master machine or central processor. After that, master machine broadcasts the aggregated parameters to node machines for new updates.  However, when the number of machines increases, the above-mentioned variable selection procedure may results in heavy computation burden and limited communication resources for master machine. Even worse, we can not address a large communication network size that the master machine can be connected to all other local machines. \par
The GADMM was first introduced by \cite{elgabli2020gadmm}. The core idea of the GADMM is to divide a group for all node machines connected with a chain into two groups(head group and tail group), then each machine in the head(or tail) group only communicates with two neighboring machines in the tail(or head) group to connect other machines except for the edge machines(first and last machines).
Without losing generality, the number of machines is considered as even $K$ in the decentralized framework. Compared to the master-worker distributed ADMM, GADMM has the advantage that it only requires ${K}/{2}$ machines to carry out computation in parallel, so it effectively reduces the communication cost required by each machine. More technical details and convergence results are discussed in the \cite{elgabli2020gadmm}. However, the GADMM algorithm can not be directly applied to computing the distributed penalized linear regression with compositional covariates in a decentralized regime due to the difficult and complex optimization problem with two different constraints and weighted $L_1$-penalized regularization problems. In the next subsection, we will solve these intractable problems systematically.\par
 \subsection{A Distributed Sparse Group Coordinate Descent Method of Multipliers}
 In this subsection, we will construct the variable selection procedure for the massive linear regression with compositional covariates in a decentralized manner, which aims to solve the issue with the whole data distributed across $K$ machines $\{\mathscr{M}_k\}_{k=1}^K$ and allow the communication of each
machine to connect only two neighbors. To derive the fast and communication-efficient algorithm for dealing with the above-mentioned procedure, we reformulate the weighted optimization problem with two kinds of constraints in \eqref{optpA} as follows
\begin{equation}\label{optpl}
\begin{split}
\widehat{\cude{\zeta}}&=\uset{\{\cude{\zeta}_k\}_{k=1}^K}{\oper{argmin}}
\sum_{k=1}^{K}\left\{\frac{1}{2n_k}\|\cude{y}_k-\cude{\Pi}_k\cude{\zeta}_k\|^2
+\lambda\|\cude{\omega}_k\circ\cude{\zeta}_k\|_1\right\},\\
&\text{subject}~\text{to}~
\begin{cases}
\cude{C}^\top\cude{\zeta}_k=0,\\
\cude{\zeta}_k=\cude{\zeta}_{k+1},
\end{cases}
k=1,\ldots, K-1.
\end{split}
\end{equation}\par
Analogously to GADMM and the literature \cite{giesen2019combining}, for the penalty parameter $\rho>0$, our extension work on the augmented Lagrangian function for problem \eqref{optpl} is given by
\begin{equation}\label{lm}
\begin{split}
\mathscr{L}_{\rho}\big(\{\cude{\zeta}_k\}^K_{k=1},&\{\mu_k\}_{K=1}^K,\{\cude{\gamma}_k\}_{k=1}^K\big)=\sum_{k=1}^K\left\{\frac{1}{2n_k}\|\cude{y}_k-\cude{\Pi}_k\cude{\zeta}_k\|^2
+\lambda\|\cude{\omega}_k\circ\cude{\zeta}_k\|_1\right\}+\sum_{k=1}^K\sum_{j=1}^d\mu_kc_j\zeta_{k,j}\\
&+\frac{\rho}{2}\sum_{k=1}^K\left(\sum_{j=1}^dc_j\zeta_{k,j}\right)^2+\sum_{k=1}^K\left\langle \cude{\gamma}_k, \cude{\zeta}_k-\cude{\zeta}_{k+1}\right\rangle
+\frac{\rho}{2}\sum_{k=1}^K\|\cude{\zeta}_k-\cude{\zeta}_{k+1}\|^2,
\end{split}
\end{equation}
where $\mu_k\in \mathbbm{R}^1$ and $\cude{\gamma}_k\in \mathbbm{R}^d$ present the Lagrangian multipliers or dual variables. Let $\mathscr{N}_h=\{\mathscr{M}_1,\mathscr{M}_3,\ldots,\mathscr{M}_{K-1}\}$ and $\mathscr{N}_t=\{\mathscr{M}_2,\mathscr{M}_4,\ldots,\mathscr{M}_{K}\}$ denote the collections of machines in head group and tail group, respectively. Correspondingly, at iteration $l+1$, the updating rules for primal variables in the head group can be written as
\begin{equation}\label{nu2}
\begin{split}
&\cude{\zeta}_k^{l+1}:=\arg\min_{\cude{\zeta}_k}\frac{1}{2n_k}\|\cude{y}_k-\cude{\Pi}_k\cude{\zeta}_k\|^2+\lambda\|\cude{\omega}_k\circ\cude{\zeta}_k\|_1\\
&~~~~~~~~~~~~~~~~~~~+\mu_k^l\sum_{j=1}^dc_j\zeta_{k,j}+\frac{\rho}{2}(\sum_{j=1}^dc_j\zeta_{k,j})^2\\
&~~~~~~~~~~~~~~~~~~~+\left\langle\cude{\gamma}_k^l,\cude{\zeta}_k-\cude{\zeta}_{k+1}^l\right\rangle+\frac{\rho}{2}\|\cude{\zeta}_k-\cude{\zeta}_{k+1}^l\|^2,~k=1.
\end{split}
\end{equation}
\begin{equation}\label{nu1}
\begin{split}
&\cude{\zeta}_k^{l+1}:=\arg\min_{\cude{\zeta}_k}\frac{1}{2n_k}\|\cude{y}_k-\cude{\Pi}_k\cude{\zeta}_k\|^2+\lambda\|\cude{\omega}_k\circ\cude{\zeta}_k\|_1+\mu_k^l\sum_{j=1}^dc_j\zeta_{k,j}
+\frac{\rho}{2}(\sum_{j=1}^dc_j\zeta_{k,j})^2\\
&~~~~~~~~~~~~~~~~~~~+\left\langle\cude{\gamma}_{k-1}^l,\cude{\zeta}_{k-1}^l-\cude{\zeta}_{k}\right\rangle+\left\langle\cude{\gamma}_{k}^l,\cude{\zeta}_{k}-\cude{\zeta}_{k+1}^l\right\rangle\\
&~~~~~~~~~~~~~~~~~~~+\frac{\rho}{2}\|\cude{\zeta}_{k-1}^l-\cude{\zeta}_{k}\|^2
+\frac{\rho}{2}\|\cude{\zeta}_{k}-\cude{\zeta}_{k+1}^l\|^2,~k\in\mathscr{N}_h\backslash \{1\}.
\end{split}
\end{equation}
\par
Send the updates $\cude{\zeta}_k^{l+1}$'s in head group to their two neighbors in tail group and carry out following updates
\begin{equation}\label{nu3}
\begin{split}
&\cude{\zeta}_k^{l+1}:=\arg\min_{\cude{\zeta}_k}\frac{1}{2n_k}\|\cude{y}_k-\cude{\Pi}_k\cude{\zeta}_k\|^2+\lambda\|\cude{\omega}_k\circ\cude{\zeta}_k\|_1+\mu_k^l\sum_{j=1}^dc_j\zeta_{k,j}
+\frac{\rho}{2}(\sum_{j=1}^dc_j\zeta_{k,j})^2\\
&~~~~~~~~~~~~~~~~~~~+\left\langle\cude{\gamma}_{k-1}^l,\cude{\zeta}_{k-1}^{l+1}-\cude{\zeta}_{k}\right\rangle+\left\langle\cude{\gamma}_{k}^l,\cude{\zeta}_{k}-\cude{\zeta}_{k+1}^{l+1}\right\rangle\\
&~~~~~~~~~~~~~~~~~~~+\frac{\rho}{2}\|\cude{\zeta}_{k-1}^{l+1}-\cude{\zeta}_{k}\|^2
+\frac{\rho}{2}\|\cude{\zeta}_{k}-\cude{\zeta}_{k+1}^{l+1}\|^2,~k\in\mathscr{N}_h\backslash \{K\}.
\end{split}
\end{equation}
\begin{equation}\label{nu4}
\begin{split}
&\cude{\zeta}_k^{l+1}:=\arg\min_{\cude{\zeta}_k}\frac{1}{2n_k}\|\cude{y}_k-\cude{\Pi}_k\cude{\zeta}_k\|^2+\lambda\|\cude{\omega}_k\circ\cude{\zeta}_k\|_1\\
&~~~~~~~~~~~~~~~~~~~+\mu_k^l\sum_{j=1}^dc_j\zeta_{k,j}+\frac{\rho}{2}(\sum_{j=1}^dc_j\zeta_{k,j})^2\\
&~~~~~~~~~~~~~~~~~~~+\left\langle\cude{\gamma}_{k-1}^l,\cude{\zeta}_{k-1}^{l+1}-\cude{\zeta}_{k}\right\rangle+\frac{\rho}{2}\|\cude{\zeta}_{k-1}^{l+1}-\cude{\zeta}_{k}\|^2,~k=K.
\end{split}
\end{equation}
After receiving the updates from neighbors, every machine locally updates its dual variables $\mu_k^{l+1}$ and $\cude{\gamma}_{k}^{l+1}$.
\begin{equation}\label{nu5}
\mu_k^{l+1}\leftarrow\mu_k^l+\rho\sum_{j=1}^dc_{j}\zeta_{k,j}^{l+1},~k=1,\ldots, K.
\end{equation}
\begin{equation}\label{nu6}
\cude{\gamma}_{k}^{l+1}\leftarrow \cude{\gamma}_{k}^{l} + \rho(\cude{\zeta}_k^{l+1}-\cude{\zeta}_{k+1}^{l+1}),~k=1,\ldots, K-1.
\end{equation}\par
To facilitate the presentation and save the space, we let $$\mathcal{A}_j^{l,b}=\frac{1}{n_k}\cude{\pi}_{k,j}^\top\left(\cude{y}_k
-\sum_{m\neq j}\zeta_{k,m}^{l,b+I(m<j)}\cude{\pi}_{k,m}\right)-\rho \left(\frac{\mu_k^{l}c_j}{\rho}+c_j\sum_{m\neq j}c_m\zeta_{k,m}^{l, b+I(m<j)}\right).$$ Accordingly, the subproblems \eqref{nu2}---\eqref{nu4} at iteration $l+1$ can be solved by the coordinate descent algorithm. That is, for $b=1, \ldots, B$, the head machines component-wisely carry out following coordinate descent updates until the convergence criterion is met,
\begin{equation}\label{U2}
\begin{split}
\cude{\zeta}_{k}^{l, b+1}\leftarrow \left[\frac{\text{Shrink}\left\{
\mathcal{A}_j^{l,b}+\rho\zeta_{k+1,j}^l-\gamma_{k,j}^l,\lambda\omega_{k, j}
\right\}}{\rho(1+c_j^2)+\frac{1}{n_k}\|\cude{\pi}_{k,j}\|^2}\right]_{1\leq j\leq d}, k=1.
\end{split}
\end{equation}
\begin{equation}\label{U1}
\begin{split}
&\cude{\zeta}_{k}^{l, b+1}\leftarrow\\
& \left[\frac{\text{Shrink}\left\{
\mathcal{A}_j^{l,b}+\rho\Big(\zeta_{k-1, j}^l+ \zeta_{k+1,j}^l\Big)+\gamma^l_{k-1,j}-\gamma_{k,j}^l,\lambda\omega_{k, j}
\right\}}{\rho(2+c_j^2)+\frac{1}{n_k}\|\cude{\pi}_{k,j}\|^2}\right]_{1\leq j \leq d}, k\in \mathscr{N}_h\backslash\{1\}.
\end{split}
\end{equation}
\par
Transmit the updated $\cude{\zeta}_k^{l+1}=\cude{\zeta}_{k}^{l,B}$ in head machines to their two neighbors in tail machines and carry out the following  coordinate descent update suntil the convergence criterion is met,
\begin{equation}\label{U3}
\begin{split}
&\cude{\zeta}_{k}^{l, b+1}\leftarrow\\
& \left[\frac{\text{Shrink}\left\{
\mathcal{A}_j^{l,b}+\rho(\zeta_{k-1, j}^{l+1}+ \zeta_{k+1,j}^{l+1}\Big)+\gamma^l_{k-1,j}-\gamma_{k,j}^l,\lambda\omega_{k, j}
\right\}}{\rho(2+c_j^2)+\frac{1}{n_k}\|\cude{\pi}_{k,j}\|^2}\right]_{1\leq j \leq d}, k\in \mathscr{N}_t\backslash\{K\}.
\end{split}
\end{equation}
\begin{equation}\label{U4}
\begin{split}
&\cude{\zeta}_{k}^{l, b+1}\leftarrow\left[\frac{\text{Shrink}\left\{
\mathcal{A}_j^{l,b}+\rho\zeta_{k-1, j}^{l+1}+\gamma^l_{k-1,j},\lambda\omega_{k, j}
\right\}}{\rho(1+c_j^2)+\frac{1}{n_k}\|\cude{\pi}_{k,j}\|^2}\right]_{1\leq j \leq d}, k=K.
\end{split}
\end{equation}\par
We summarize the  \textbf{d}istributed \textbf{s}parse \textbf{g}roup \textbf{c}oordinate \textbf{d}escent \textbf{m}ethod of \textbf{m}ultipliers\\(DSGCDMM) for solving large-scale penalized linear regression with compositional covariates in Algorithm \ref{alg2}.\par
\begin{algorithm}[ht!]
 \renewcommand{\algorithmicrequire}{\textbf{Input:}}
 \renewcommand{\algorithmicensure}{\textbf{Output:}}
\caption{DSGCDMM: Distributed sparse group coordinate descent method of multipliers for solving large-scale penalized linear regression with compositional covariates.}
\label{alg2}
\begin{algorithmic}[1]
\REQUIRE  data $\{\cude{y}_{k}, \cude{\Pi}_{k}\}_{k=1}^K$ on machines $\{\mathscr{M}_k\}_{k=1}^K$, the value of constant $\rho$, the number of rounds of communication $L$.
\ENSURE $\cude{\zeta}^L_K$.
\STATE {\textbf{Initialization}:
\begin{itemize}
 \item $\mathscr{N}_h=\{\cude{\zeta}_k\mid k: odd\}$, $\mathscr{N}_t=\{\cude{\zeta}_k\mid k: even\}$.\par
 \item $\cude{\zeta}_k^{(0)}=\cude{\gamma}_k^{(0)}=\cude{0}$, $\mu_k^{(0)}=0$ for all $k$.
 \item Calculate $\cude{\omega}_k$ in the local machines.
 \end{itemize}
 }
\FOR{$l=0,1,2,\ldots, L-1$}
\FOR{Head machines $\{\mathscr{M}_k: k\in \mathscr{N}_h\}$}
\STATE Carry out the coordinate descent Steps in master machine until the convergence criterion is met.
\STATE Initialize $\cude{\zeta}_{k}^{l,0}=\cude{\zeta}_{k}^l$.
\FOR{$b=0,1,2,\ldots$}
\FOR{$j=1,\ldots,d$}
\STATE Update $\zeta_{k,j}^{l, b+1}$ via \eqref{U2} and \eqref{U1} in parallel.
\ENDFOR
\STATE Set $\cude{\zeta}_{k}^{l+1}=\cude{\zeta}_{k}^{l,B}$. Send $\cude{\zeta}_{k}^{l+1}$ to its tail neighboring machines $\mathscr{M}_{k-1}$ and $\mathscr{M}_{k+1}$.
\ENDFOR
\ENDFOR
\FOR{Tail machines $\{\mathscr{M}_{k}: k\in \mathscr{N}_t\}$}
\STATE Carry out the coordinate descent Steps in master machine until the convergence criterion is met.
\STATE Initialize $\cude{\zeta}_{k}^{l,0}=\cude{\zeta}_{k}^l$.
\FOR{$b=0,1,2,\ldots$}
\FOR{$j=1,\ldots,d$}
\STATE Update $\zeta_{k,j}^{l, b+1}$ via \eqref{U3} and \eqref{U4} in parallel.
\ENDFOR
\STATE Set $\cude{\zeta}_{k}^{l+1}=\cude{\zeta}_{k}^{l,B}$. Send $\cude{\zeta}_{k}^{l+1}$ to its head neighboring machines $\mathscr{M}_{k-1}$ and $\mathscr{M}_{k+1}$.
\ENDFOR
\ENDFOR
\STATE Every machine locally updates the dual variables $\mu_k^{l+1}$, $\cude{\gamma}_{k-1}^{l+1}$ and $\cude{\gamma}_{k}^{l+1}$ via \eqref{nu5} and \eqref{nu6}.
\ENDFOR
\end{algorithmic}
\end{algorithm}\par
Remark that, for the LASSO and adaptive LASSO penalties, we seek to facilitate the calculation by obtaining the $\cude{\omega}$, $\cude{\omega}_k$ in algorithm \ref{alg1} and algorithm \ref{alg2} from $1st$ machine $\mathscr{M}_1$  with data set $\{\cude{y}_1, \cude{\Pi}_1\}$ and broadcasting it to other machines.
\subsection{Regularization Parameter Selection}\label{RPS}
For model \eqref{optpA}, \eqref{optSCAD} and our distributed algorithm \ref{alg1} and algorithm \ref{alg2}, the regularization parameter $\lambda$ needs to be specified. In this article,  we adopt the generalized information criterion(GIC, \cite{fan2013tuning}) to select an optimal regularization parameter $\lambda_{opt}$. The GIC is also modified by \cite{lin2014variable} that aims to obtain the optimal regularized solution path for penalized linear regression with compositional covariates. We adopt the subdata set at first machine $\mathscr{M}_1$ to determine $\lambda_{opt}$. The GIC is defined as
\begin{equation}
\text{GIC}(\lambda):=\log\left\{\frac{\|\cude{y}_1-\cude{\Pi}_1\cude{\widehat{\zeta}}(\lambda)\|^2}{n_1}\right\}
+(\text{df}_\lambda-1)\frac{\log\log n_1}{n_1}\log(d\vee n_1),
\end{equation}
where $\cude{\widehat{\zeta}}(\lambda)$ is the regularized estimator, $\text{df}_\lambda-1$ is the effective number of non-zero parameters with respect to $\cude{\widehat{\zeta}}(\lambda)$ due to the zero-sum constraint and $\text{df}_\lambda$($\geq 2$) is the degree of freedom of the candidate model, $d\vee n_1=\max(d,n_1)$. Then, $\lambda_{opt}:=\arg\min_{\lambda}\text{GIC}(\lambda)$. We now choose a grid of values $0\leq\lambda_{\min}=\lambda_1<\lambda_2<\cdots<\lambda_S=\lambda_{\max}$ for $\lambda$. Let $\cude{\Pi}_1=(\cude{\pi}_{1,1},\ldots,\cude{\pi}_{1,p})$. As recommended in section 2.12.1 and section 9.2.4 of \cite{buhlmann2011statistics},
we determine $$\lambda_{\min}\thickapprox \frac{1}{n_1},$$ $$\lambda_{\max}=\min\left\{\max_{1\leq j\leq p}\left|\frac{2\cude{\pi}^\top_{1,j}\cude{y}_1}{n_1}\right|, \max_{1\leq j\leq p}\left|\frac{2\cude{\pi}^\top_{1,j}\cude{y}_1}{\sqrt{n_1}\|\cude{y}_1\|}\right|\right\},$$
and $$ \lambda_{s+1}=\lambda_s \exp\Delta,~s=1,\ldots,S-1,$$
with
$$\Delta=\frac{\log(\lambda_{\max})-\log(\lambda_{\min})}{S-1}.$$
\section{Convergence Analysis}\label{CT}
In this section, we turn to statements of our main theoretical results on the proposed distributed computing approaches. Because the convergence theories of ADMM-based algorithm DSCDMM with two types of constraints have been investigated by \cite{mateos2010distributed,giesen2016distributed,giesen2019combining}, this work only discusses the convergence of DSGCDMM algorithm. In other words, we seek to prove that the update of each $k$, $\cude{\zeta}_k^{l+1}$ at iteration $l+1$, which is calculated by Algorithm \ref{alg2}, converges to the optimal solution(denoted as $\cude{\zeta}_k^\star$) of the problem in \eqref{optpl} as $l\to\infty$. To facilitate notation, we let $\mathbb{Q}_k(\cude{\zeta}_k):=\frac{1}{2n_k}\|\cude{y}_k-\cude{\Pi}_k\cude{\zeta}_k\|^2
+\lambda\|\cude{\omega}_k\circ\cude{\zeta}_k\|_1$. In such setting, both parts of function $\mathbb{Q}_k(\cude{\zeta}_k)$ are closed, proper, and convex. It is worth mentioning that there exist the extra equality constraints $\cude{C}^\top\cude{\zeta}_k=0(k=1,\ldots, K)$. Following \cite{boyd2011distributed,elgabli2020gadmm}, we present primal feasibility and dual feasibility as the necessary and sufficient optimality conditions for problem \eqref{optpl}, as follows:
\begin{itemize}
    \item{Primal Feasibility} 
    \begin{equation}\label{PF}
        \begin{split}
            &\cude{\zeta}_k^\star=\cude{\zeta}_{k-1}^\star,~k=2,\ldots,K,\\
            &\cude{C}^\top\cude{\zeta}^\star=0,~k=1,\ldots,K.
        \end{split}
    \end{equation} 
    \item{Dual Feasibility} 
    \begin{equation}\label{DF}
    \begin{split}
         \cude{0}&\in \partial\mathbb{Q}_k(\cude{\zeta}_k^\star)+\mu_k^\star\cude{C}-\cude{\gamma}_{k-1}^\star+\cude{\gamma}_k^\star,~k=2,\ldots, K-1,\\
         \cude{0}&\in \partial\mathbb{Q}_k(\cude{\zeta}_k^\star)+\mu_k^\star\cude{C}+\cude{\gamma}_k^\star,~k=1,\\
         \cude{0}&\in \partial\mathbb{Q}_k(\cude{\zeta}_k^\star)+\mu_k^\star\cude{C}+\cude{\gamma}_{k-1}^\star,~k=K.
    \end{split}
    \end{equation}
\end{itemize}
Here, $\partial$ denotes the subdifferential operator, $(\cude{\zeta}^\star,\{\mu_k^\star\}_{k=1}^K, \{\cude{\gamma}_k^\star\}_{k=1}^K)$ with $\cude{\zeta}^\star=\cude{\zeta}^\star_k=\cude{\zeta}^\star_{k-1}$(for all $k$) is the saddle point of $\mathscr{L}_{0}\left(\{\cude{\zeta}_k\}^K_{k=1},\{\mu_k\}_{k=1}^K,\{\cude{\gamma}_k\}_{k=1}^K\right)$. For all $k\in \mathscr{N}_t$, we obtain the $\cude{\zeta}^{l+1}_{k\in \mathscr{N}_t\backslash\{K\}}$ and $\cude{\zeta}^{l+1}_K$ by solving the subproblems \eqref{nu3} and \eqref{nu4} at iteration $l+1$, respectively. Then, it is follows that
\begin{equation}
\begin{split}
   \cude{0}\in &\partial \mathbb{Q}_k(\cude{\zeta}_k^{l+1})+\left(\mu_k^l+\rho\cude{C}^\top\cude{\zeta}_k^{l+1}\right)\cude{C}-\left(\cude{\gamma}_{k-1}^l+\rho(\cude{\zeta}_{k-1}^{l+1}-\cude{\zeta}_{k}^{l+1})\right)\\
   &+\left(\cude{\gamma}_k^l+\rho(\cude{\zeta}_k^{l+1}-\cude{\zeta}_{k+1}^{l+1})\right),k\in \mathscr{N}_t\backslash\{K\}.  
\end{split}
\end{equation}
\begin{equation}
    \cude{0}\in  \partial \mathbb{Q}_k(\cude{\zeta}_k^{l+1})+\left(\mu_k^l+\rho\cude{C}^\top\cude{\zeta}_k^{l+1}\right)\cude{C}-\left(\cude{\gamma}_{k-1}^l+\rho(\cude{\zeta}_{k-1}^{l+1}-\cude{\zeta}_{k}^{l+1})\right),~k=K.
\end{equation}
By the updates in \eqref{nu5} and \eqref{nu6}, we have 
\begin{equation}\label{nts1}
    \cude{0}\in \partial \mathbb{Q}_k(\cude{\zeta}_k^{l+1})+\mu_k^{l+1}\cude{C}-\cude{\gamma}_{k-1}^{l+1}+\cude{\gamma}_{k}^{l+1},~k\in \mathscr{N}_t\backslash\{K\}.
\end{equation}
\begin{equation}\label{nts2}
     \cude{0}\in \partial \mathbb{Q}_k(\cude{\zeta}_k^{l+1})+\mu_k^{l+1}\cude{C}-\cude{\gamma}_{k-1}^{l+1},~k=K.
\end{equation}
The results in \eqref{nts1} and \eqref{nts2} indicate that the dual feasibility conditions in \eqref{DF} always holds for all $k\in \mathscr{N}_t$, and the dual residual of tail machines is always zero.\par
Similarly, for all $k\in \mathscr{N}_h$, $\cude{\zeta}^{l+1}_{k=1}$ and $\cude{\zeta}^{l+1}_{k\in \mathscr{N}_h\backslash\{1\}}$  minimizes the subproblems \eqref{nu2} and \eqref{nu1}, respectively. Correspondingly, from the first order optimality condition and by adding the terms $\cude{\zeta}_{k-1}^{l+1}$ and $\cude{\zeta}_{k+1}^{l+1}$, we have
\begin{equation}
      \cude{0}\in \partial \mathbb{Q}_k(\cude{\zeta}_k^{l+1}) + \left(\mu_k^l+\rho\cude{C}^\top\cude{\zeta}_k^{l+1}\right)\cude{C}+\left(\cude{\gamma}_k^l+\rho(\cude{\zeta}_k^{l+1}-\cude{\zeta}_{k+1}^{l+1})\right)+\rho(\cude{\zeta}_{k+1}^{l+1}-\cude{\zeta}_{k+1}^l),~k=1.
\end{equation}
\begin{equation}
\begin{split}
    \cude{0}\in &\partial \mathbb{Q}_k(\cude{\zeta}_k^{l+1}) + \left(\mu_k^l+\rho\cude{C}^\top\cude{\zeta}_k^{l+1}\right)\cude{C}-\left(\cude{\gamma}_{k-1}^l+\rho(\cude{\zeta}_{k-1}^{l+1}-\cude{\zeta}_k^{l+1})\right)\\
    &+\left(\cude{\gamma}_k^l+\rho(\cude{\zeta}_k^{l+1}-\cude{\zeta}_{k+1}^{l+1})\right)+\rho(\cude{\zeta}_{k-1}^{l+1}-\cude{\zeta}_{k-1}^l)+\rho(\cude{\zeta}_{k+1}^{l+1}-\cude{\zeta}_{k+1}^l),~k\in \mathscr{N}_h\backslash\{1\}.
    \end{split}
\end{equation}
By the updates in \eqref{nu5} and \eqref{nu6}, following formulations hold,
\begin{equation}\label{ph1}
      \cude{0}\in \partial \mathbb{Q}_k(\cude{\zeta}_k^{l+1}) + \mu_{k}^{l+1}\cude{C}+\cude{\gamma}_k^{l+1}+\rho(\cude{\zeta}_{k+1}^{l+1}-\cude{\zeta}_{k+1}^l),~k=1.
\end{equation}
\begin{equation}\label{ph2}
\begin{split}
    \cude{0}\in &\partial \mathbb{Q}_k(\cude{\zeta}_k^{l+1}) + \mu_{k}^{l+1}\cude{C}-\cude{\gamma}_{k-1}^{l+1}+\cude{\gamma}_k^{l+1}+\rho(\cude{\zeta}_{k-1}^{l+1}-\cude{\zeta}_{k-1}^l)+\rho(\cude{\zeta}_{k+1}^{l+1}-\cude{\zeta}_{k+1}^l),~k\in \mathscr{N}_h\backslash\{1\}.
    \end{split}
\end{equation}
Therefore, at iteration $l+1$, the dual residual of machine which belongs to $\mathscr{N}_h$ is defined as
\begin{equation}
\cude{s}_k^{l+1}=\left\{\begin{array}{l}
\rho\left(\boldsymbol{\zeta}_{k-1}^{l+1}-\boldsymbol{\zeta}_{k-1}^l\right)+\rho\left(\boldsymbol{\zeta}_{k+1}^{l+1}-\boldsymbol{\zeta}_{k+1}^l\right), k \in \mathscr{N}_h \backslash\{1\}, \\
\rho\left(\boldsymbol{\zeta}_{k+1}^{l+1}-\boldsymbol{\zeta}_{k+1}^l\right),  k=1,
\end{array}\right.
\end{equation}
which is the same as the dual residual in \cite{elgabli2020gadmm}.\par
Showing that the conditions in \eqref{PF}---\eqref{DF} are satisfied for each machine $\mathscr{M}_k$ is necessary to demonstrate the convergence of DSGCDMM. Let $\cude{r}_{k}^{l+1}=\cude{\zeta}_k^{l+1}-\cude{\zeta}_{k+1}^{l+1}(k=1,\ldots,K-1)$ and $g_k^{l+1}=\cude{C}^\top\cude{\zeta}_k^{l+1}(k=1,\ldots,K)$ be the primal residuals of the  constraints $\cude{\zeta}_k=\cude{\zeta}_{k+1}$ and $\cude{C}^\top\cude{\zeta}_k=0$, respectively. Since the dual residual of tail machines is always zero, it is sufficient to show that $\cude{r}_k^{l}$ for all $k=1,\ldots,K-1$, $g_k^{l}$ for all $k=1,\ldots,K$ and $\cude{s}_{k\in\mathscr{N}_h}^{l}$ convenge to zero, and $\sum_{k=1}^K\mathbb{Q}_k(\cude{\zeta}_k^l)$ converges to $\sum_{k=1}^K\mathbb{Q}_k(\cude{\zeta}_k^\star)$ as $l\to\infty$. To derive the convergence results for DSGCDMM, we present Lemma below as the first technical result in our work.\par
\begin{lem}\label{lem1}
    For all $k=1,\ldots,K$, given $\lambda>0$, $\rho>0$, when the updates $\cude{\zeta}_k^{l+1}$'s are calculated by carrying out Algorithm \ref{alg2}, then the optimality gap is upper bounded as 
    \begin{equation}
\sum_{k=1}^K\left(\mathbb{Q}_k(\cude{\zeta}_k^{l+1})-\mathbb{Q}_k(\cude{\zeta}_k^{\star})\right)\leq -\sum_{k=1}^K\mu_k^{l+1}g_k^{l+1}-\sum_{k=1}^{K-1}\left\langle\cude{\gamma}_k^{l+1},\cude{r}_k^{l+1}\right\rangle+\sum_{k\in\mathscr{N}_h}\left\langle\cude{s}^{l+1}_k, \cude{\zeta}_k^\star-\cude{\zeta}_k^{l+1}\right\rangle,
    \end{equation}
    and lower bounded as 
    \begin{equation}
\sum_{k=1}^K\left(\mathbb{Q}_k(\cude{\zeta}_k^{l+1})-\mathbb{Q}_k(\cude{\zeta}_k^{\star})\right)\geq -\sum_{k=1}^K\mu_k^\star g_k^{l+1}-\sum_{k=1}^{K-1}\left\langle\cude{\gamma}_{k}^\star,\cude{r}_k^{l+1}\right\rangle.
    \end{equation}
\end{lem}
Now our main technical convergence results are summarized in the following theorem under the \textbf{Lemma \ref{lem1}}.
.
\begin{thm}\label{thm1}
    Suppose that the Lagrangian $\mathscr{L}_0$ has a saddle point, i.e., $(\cude{\zeta}^\star,\{\mu_k^\star\}_{k=1}^K, \{\cude{\gamma}_k^\star\}_{k=1}^K)$. The parameter iterates $\cude{\zeta}_k^{l+1}$'s at iteration $l+1$ are obtained by running Algorithm \ref{alg2}. Then, as $l\to\infty$, 
    \begin{enumerate}
        \item the primal residual of the  constraint $\cude{C}^\top\cude{\zeta}_k=0$ converges to zero, i.e., \begin{equation}
    g_k^l\rightarrow 0,~k=1,\ldots,K,
        \end{equation}
        \item the primal residual of the  constraint $\cude{\zeta}_k=\cude{\zeta}_{k+1}$  converges to zero, i.e.,
         \begin{equation}
    \cude{r}_k^l\rightarrow\cude{0},~k=1,\ldots,K-1,
        \end{equation}
        \item the dual residual converges to zero, i.e.,
        \begin{equation}
            \cude{s}_k^l\rightarrow \cude{0},~k\in\mathscr{N}_h,
        \end{equation}
        \item the optimality gap converges to zero, i.e.,
        \begin{equation}
            \sum_{k=1}^K\left(\mathbb{Q}_k(\cude{\zeta}_k^{l})-\mathbb{Q}_k(\cude{\zeta}_k^{\star})\right)\rightarrow 0.
        \end{equation}
    \end{enumerate}
\end{thm}
The detailed proof of \textbf{Lemma \ref{lem1}} and \textbf{Theorem \ref{thm1}} are given in the Appendix \ref{app1} and Appendix \ref{app2}.
\begin{remark}
    It is worth noting that \cite{mateos2010distributed} proposed the distributed LASSO algorithms based on the ADMM method applicable in situations where the training data are distributed across different machines(agents) and they are unable to communicate with a central processing unit. Furthermore, they proved that all the estimates of the local parsmeters calculated by their developed distributed LASSO algorithms(i.e., distributed quadratic
programming (DQP-)LASSO algorithm, distributed coordinate descent (DCD-)LASSO algorithm, and Distributed (D-)LASSO) converge to the global LASSO solution if the associated distributed system is a connected graph. Although we proved the convergence of the algorithm DSGCDMM, it still remains
the convergence theories to be investigated systemically that our distributed penalized estimates obtained by DSGCDMM perform as well as the global estimates when the number of rounds of communication goes to infinity. In addition, the statistical asymptotic properties such as consistency and asymptotic normality under some regularity conditions need to be further studied in the future.
\end{remark}
\section{Numerical Experiments}\label{NET}
In this section, we numerically investigate the finite-sample statistical performance of our proposed algorithms with   DSCDMM and  DSGCDMM by using synthetic data and real data. All the numerical studies are conducted on a server with Intel(R) Xeon(R) Gold 6230R CPU @ 2.10GHz Processors and 64 Gbytes RAM running Windows 10, on which we use the software R, version 4.2.3. All the simulation results are based on 100 independent replications, and are calculated on a single machine. Therefore, we only consider the statistical efficiency and neglect the computational efficiency.
\subsection{Synthetic Data}\label{NE}
We investigate a popular simulation regression model with compositional covariates from \cite{lin2014variable} to generate the synthetic data for comparisons. In particular, for the part of compositional covariates, let $\cude{\Delta} =(\delta_{ij})$ be an $n\times p$ data matrix that is generated from a  multivariate normal distribution $\cude{N}_p(\cude{\nu}, \cude{\Sigma})$. Then the covariate matrix $\cude{X}=(x_{ij})$  is obtained by the transformation $x_{ij}= \exp(\delta_{ij})/\sum_{l=1}^p\exp(\delta_{il})$. The non-compositional covariates $\cude{V}$$'s$ for different examples are generated from the several multivariate distributions. Consequently, the response $\cude{Y}$ is generated by the formulation \eqref{cn}. Here, the true coefficient vector $\cude{\zeta}_0=(\cude{\beta}_0^\top, \cude{\theta}_0^\top)^\top$ incuding compositional and non-compositional covariates  are respectively  considered as
$$\cude{\beta}_0=(1, -0.8, 0.6, 0, 0, -1.5, -0.5, 1.2, \cude{0}_{p-8}^\top)^\top,$$
$$\cude{\theta}_0=(0.7, -1.5, 1, 0, 0, 0, -0.8, 2.3, \cude{0}_{q-8}^\top)^\top,$$
where $\cude{0}_{p(q)-8}=(0,\ldots,0)^\top\in\mathbbm{R}^{p(q)-8}$. To evaluate the performance of the distributed variable selection methods for the compositional data, four algorithms with two various penalties, i.e. adaptive LASSO(AL) and SCAD(S), are considered to draw comparisons. Four algorithms are given below, (1) the centralized algorithm  DSCDMM(denoted as DSC-AL and DSC-S) introduced in Section \ref{sec3}; (2) the decentralized algorithm  DSGCDMM(DSCG-AL and DSCG-S for short) presented in Section \ref{DA}; (3) the global coordinate descent method of multipliers(GCDMM)(GC-AL and GC-S for short) from \cite{lin2014variable}; and (4) the averaging-based estimator based on the local estimators calculated by coordinate descent method of multipliers(ACDMM)(labeled as AC-AL, AC-S), which is an one-shot method that requires only one
round of communication.
 As introduced in Subection \ref{RPS}, the optimal regularization parameter $\lambda$ is selected by the GIC. The penalty parameter $\rho$ is chosen as $10^{-3}$. We take the total sample size $n = 200,000$ and the number of machines $K \in \{10, 100, 200\}$. For simplicity, we set the sample size $n_k$ for $k_{th}$ local machine as $n_k=\lfloor n/K\rfloor$, where $\lfloor c\rfloor$ denotes the the largest integer which is not greater than the constant $c$. What is more, we adopt several performance measures for our comparisons to evaluate the effectiveness. Namely, the estimation accuracy is measured by the average estimation error(AEE), i.e. $\text{AEE}=100^{-1}\sum_{s=1}^{100}\|\cude{\widehat{\zeta}}^{(s)}-\cude{\zeta}_0\|$. The variable selection accuracy is assessed by the average number of false positives(FP) and the average number of false negatives(FN), where positives and negatives stand for nonzero and zero coefficients, respectively. Furthermore, we report the variable selection accuracy results including FPs and FNs for the compositional(C) and non-compositional(NC) covariates, which are labeled as FP-C, FP-NC, FN-C, FN-NC.
We set $\cude{\nu}$, $\cude{\Sigma}$, $p$, $q$, $\cude{V}$ and error term $\cude{\epsilon}$ from different scenarios in the following simulation experiment. \par
 Let the component $\nu_j$ of $\cude{\nu}$ be
\begin{align*}
\nu_j=
\begin{cases}
\log(0.2p),& j=1,\ldots, 5,\\
0,& \text{otherwise,}
\end{cases}
\end{align*}
and $\cude{\Sigma}=(\sigma^{|i-j|})$ with $\sigma\in\{0.2, 0.5\}$. The dimensions $p$ and $q$ are fixed as $p=15$ and $q=10$, respectively. Then, $d=25$. The matrix for non-compositional covariates $\cude{V}$ is generated form following distributions:
\begin{itemize}
\item{\textbf{Case 1}} Heavy-tailed distribution: $\cude{T}_3(\cude{0}_q, \cude{\Xi})$, where $\Xi_{ij}=0.5^{i\neq j}$.
\item{\textbf{Case 2}}  Right-skewed distribution: $\cude{LN}(\cude{0}_q, \cude{\Xi})$, where $\Xi_{ij}=0.5^{|i - j|}$.
\end{itemize}
 The error term $\cude{\epsilon}$ is generated from $N(0, 0.2^2)$.
\par
Tables \ref{reAL1}---\ref{reS2} summarize all the simulation results for variable selection accuracy when carrying out different algorithms with three types of penalties for \textbf{Case 1.} and \textbf{Case 2.}. We see from the tables that, regardless of the values of $\rho'$s and $K'$s, most of the false positives and false negatives for DSCDMM, DSGCDMM, and GCDMM are zero, except for ACDMM. This indicates that the DSGCDMM algorithm works well when handling variable selection problems with componsitional covariates. However, in contrast to DSGCDMM, ACDMM performs poorly when dealing with variable selection problems in a distributed system.\par
\begin{figure}[htbp]
\centering
\includegraphics[width=16cm,height=16cm]{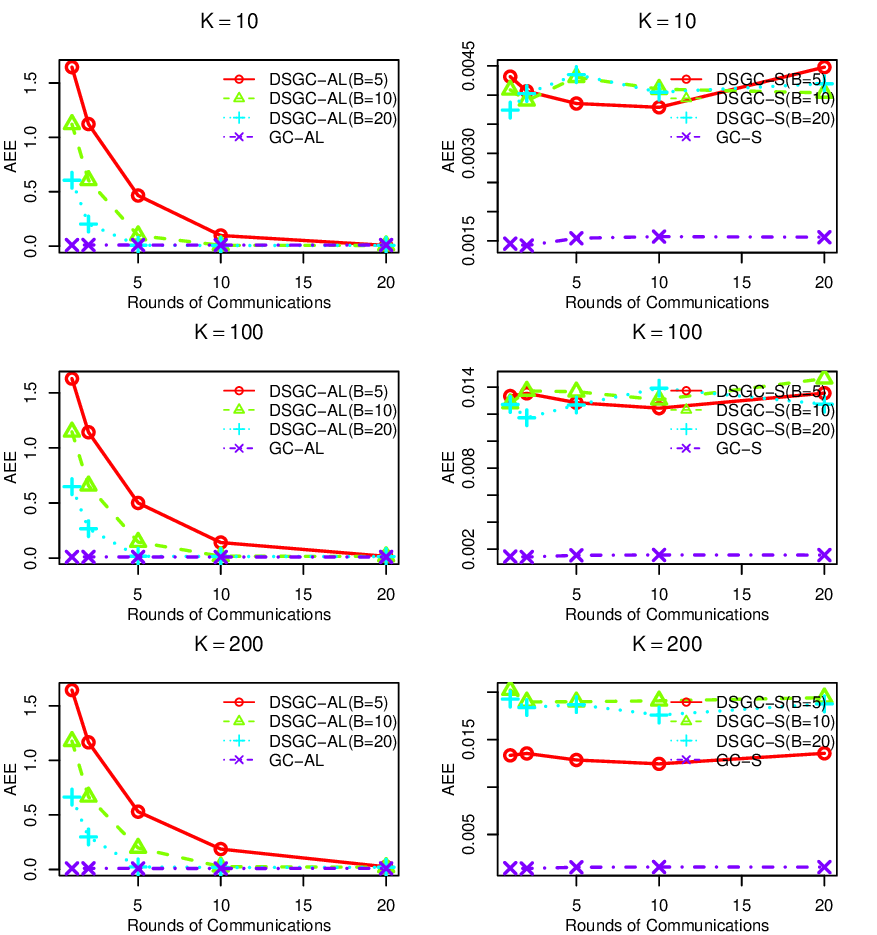}
\caption{The AEEs versus rounds of communications for \textbf{Case.1} based on DSGCDMM algorithm against different  $B'$s, where $\sigma=0.2$. The  columns 1---2 are for the results based on different penalties. \label{GC112}}
\end{figure}
\begin{figure}[htbp]
\centering
\includegraphics[width=16cm,height=16cm]{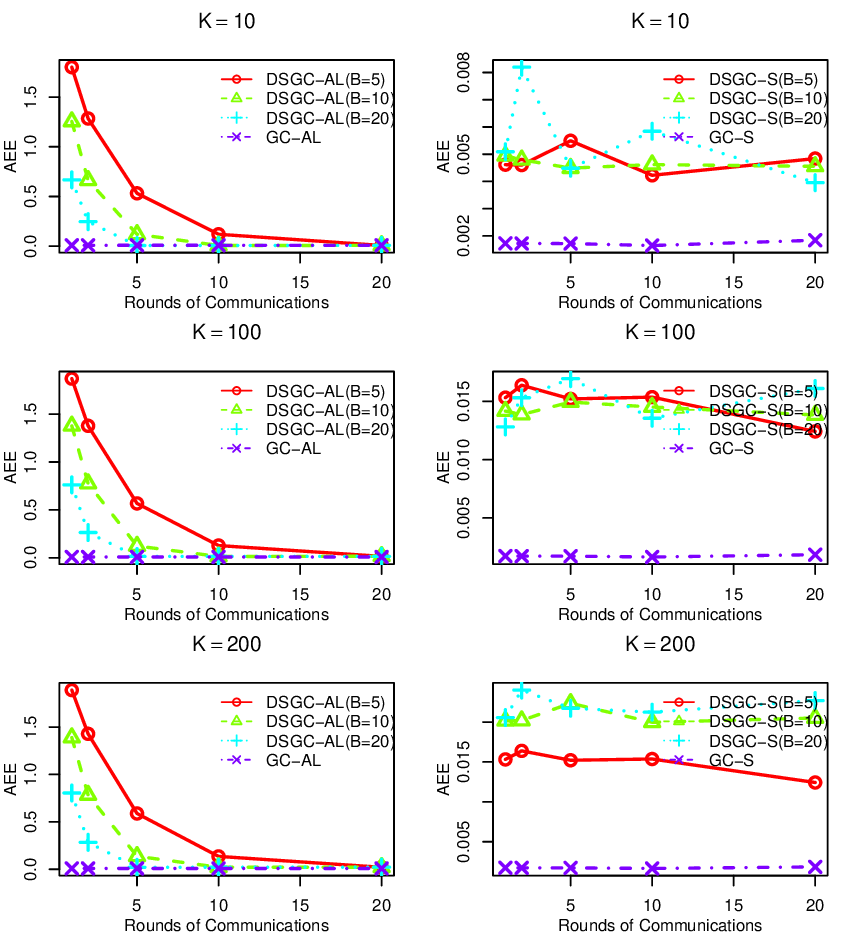}
\caption{The AEEs versus rounds of communications for \textbf{Case.1} based on DSGCDMM algorithm against different  $B'$s, where $\sigma=0.5$. The  columns 1---2 are for the results based on different penalties. \label{GC115}}
\end{figure}

To further validate the estimation accuracy of our proposed DSGCDMM algorithm, we compare the AEEs by setting three different values of the maximum number of iterations (denoted as $B$) in Algorithm  \ref{alg2} under different penalty types. Specifically, we consider $B \in \{5, 10, 20 \}$ and five different values of the number of communication rounds, denoted by $L \in \{1, 2, 5, 10, 20\}$. Figures \ref{GC112}---\ref{GC115} report the AEEs for \textbf{Case 1.} and \textbf{Case 2.} under different simulation settings.\par
From the simulation results, we can observe that all the AEEs of DSGC-AL decrease as $L$ increases to 20.  Furthermore, we observe that the AEEs at $B=20$ converge rapidly to those of GC-AL, and are smaller than those at $B=5$ and $B=10$. This implies that the proposed approach, DSGCDMM with adaptive LASSO penalty, yields an estimator that performs as well as the global method, GCDMM with adaptive LASSO penalty, as the number of communication rounds approaches infinity. The results shown in the figures intuitively demonstrate that the estimation accuracy of DSGC-AL increases as the number of communication rounds and the maximum number of iterations in Algorithm \ref{alg2} increase. Furthermore, we observe that the AEEs of DSGC-S are consistently very close to those of GC-S, regardless of the values of $L'$s and $B'$s. Furthermore, we observe that the AEEs of DSGC-S are significantly smaller than those of DSGC-AL. This is because SCAD-based approaches require the LLA/CD algorithm, which results in our DSGCDMM algorithm with the SCAD penalty type converging quickly. To conclude, the estimator obtained by implementing the DSGCDMM algorithm can achieve comparable performance with the global estimator, provided that an appropriate number of communication rounds and maximum number of iterations are used.\par
\begin{landscape}
\begin{table}[!htbp] \centering
  \caption{Variable selection performance for \textbf{Case 1.} when the penalty function is determined as AL type.}
  \label{reAL1}
\begin{tabular}{@{\extracolsep{5pt}} ccccccccc}
\\[-1.8ex]\hline
\hline \\[-1.8ex]
&Number of Machines&Method & FP& FN & FP-C & FP-NC & FN-C & FN-NC \\
\hline \\[-1.8ex]
\multirowcell{14}{$\sigma=0.2$}&\multirowcell{4}{$K=10$}&DSGC-AL &  0.0(0.000) & 0(0) & 0.00(0.000) & 0.00(0.000) & 0(0) & 0(0) \\
&&DSC-AL &    0.0(0.000) & 0(0) & 0.00(0.000) & 0.00(0.000) & 0(0) & 0(0) \\
&&GC-AL &   0.0(0.000) & 0(0) & 0.00(0.000) & 0.00(0.000) & 0(0) & 0(0) \\
&&AC-AL &  0.0(0.000) & 0(0) & 0.00(0.000) & 0.00(0.000) & 0(0) & 0(0) \\
\\
&\multirowcell{4}{$K=100$}&DSGC-AL &  \textbf{0.0(0.000)} & 0(0) & 0.00(0.000) & 0.00(0.000) & 0(0) & 0(0) \\
&&DSC-AL &  0.0(0.000) & 0(0) & 0.00(0.000) & 0.00(0.000) & 0(0) & 0(0) \\
&&GC-AL &   0.0(0.000) & 0(0) & 0.00(0.000) & 0.00(0.000) & 0(0) & 0(0) \\
&&AC-AL &   \textbf{0.2(0.523)} & 0(0) & 0.15(0.489) & 0.05(0.224) & 0(0) & 0(0) \\
\\
&\multirowcell{4}{$K=200$}&DSGC-AL &  \textbf{0.0(0.000)} & 0(0) & 0.00(0.000) & 0.00(0.000) & 0(0) & 0(0) \\
&&DSC-AL &  0.0(0.000) & 0(0) & 0.00(0.000) & 0.00(0.000) & 0(0) & 0(0) \\
&&GC-AL &  0.0(0.000) & 0(0) & 0.00(0.000) & 0.00(0.000) & 0(0) & 0(0) \\
&& AC-AL &  \textbf{10.9(1.373)} & 0(0) & 7.40(1.273) & 3.50(1.051) & 0(0) & 0(0) \\ [-1.8ex] \\
\hline \\[-1.8ex]
\multirowcell{14}{$\sigma=0.5$}&\multirowcell{4}{$K=10$}&DSGC-AL &0.00(0.000) & 0(0) & 0.00(0.000) & 0.00(0.000) & 0(0) & 0(0) \\
&&DSC-AL &  0.00(0.000) & 0(0) & 0.00(0.000) & 0.00(0.000) & 0(0) & 0(0) \\
&&GC-AL &  0.00(0.000) & 0(0) & 0.00(0.000) & 0.00(0.000) & 0(0) & 0(0) \\
&&AC-AL &  0.00(0.000) & 0(0) & 0.00(0.000) & 0.00(0.000) & 0(0) & 0(0) \\
\\
&\multirowcell{4}{$K=100$}&DSGC-AL & \textbf{0.00(0.000)} & 0(0) & 0.00(0.000) & 0.00(0.000) & 0(0) & 0(0) \\
&&DSC-AL &    0.00(0.000) & 0(0) & 0.00(0.000) & 0.00(0.000) & 0(0) & 0(0) \\
&&GC-AL &   0.00(0.000) & 0(0) & 0.00(0.000) & 0.00(0.000) & 0(0) & 0(0) \\
&&AC-AL &   \textbf{4.95(6.621)} & 0(0) & 3.25(4.253) & 1.70(2.386) & 0(0) & 0(0) \\
\\
&\multirowcell{4}{$K=200$}&DSGC-AL &  \textbf{0.25(0.444)} & 0(0) & 0.20(0.410) & 0.05(0.224) & 0(0) & 0(0) \\
&&DSC-AL &    0.00(0.000) & 0(0) & 0.00(0.000) & 0.00(0.000) & 0(0) & 0(0) \\
&&GC-AL &  0.00(0.000) & 0(0) & 0.00(0.000) & 0.00(0.000) & 0(0) & 0(0) \\
&&AC-AL &  \textbf{12.20(1.765)} & 0(0) & 8.10(0.968) & 4.10(0.912) & 0(0) & 0(0) \\
\hline\hline \\[-1.8ex]
\multicolumn{9}{l}{Note: The respective standard errors are recorded in the parentheses. }
\end{tabular}
\end{table}
\end{landscape}
\begin{landscape}
\begin{table}[!htbp] \centering
  \caption{Variable selection performance for \textbf{Case 1.} when the penalty function is determined as SCAD type.}
  \label{reS1}
\begin{tabular}{@{\extracolsep{5pt}} ccccccccc}
\\[-1.8ex]\hline\hline \\[-1.8ex]
&Number of Machines&Method & FP& FN & FP-C & FP-NC & FN-C & FN-NC \\
\hline \\[-1.8ex]
\multirowcell{14}{$\sigma=0.2$}&\multirowcell{4}{$K=10$}&DSGC-S & 0.00(0.000) & 0(0) & 0.00(0.000) & 0.00(0.000) & 0(0) & 0(0) \\
&&DSC-S &   0.00(0.000) & 0(0) & 0.00(0.000) & 0.00(0.000) & 0(0) & 0(0) \\
&&GC-S &   0.00(0.000) & 0(0) & 0.00(0.000) & 0.00(0.000) & 0(0) & 0(0) \\
&&AC-S &   0.00(0.000) & 0(0) & 0.00(0.000) & 0.00(0.000) & 0(0) & 0(0) \\
\\
&\multirowcell{4}{$K=100$}&DSGC-S  & \textbf{0.00(0.000)} & 0(0) & 0.00(0.000) & 0.00(0.000) & 0(0) & 0(0) \\
&&DSC-S &   0.00(0.000) & 0(0) & 0.00(0.000) & 0.00(0.000) & 0(0) & 0(0) \\
&&GC-S &   0.00(0.000) & 0(0) & 0.00(0.000) & 0.00(0.000) & 0(0) & 0(0) \\
&&AC-S &  \textbf{1.75(3.582)} & 0(0) & 1.50(2.911) & 0.25(0.910) & 0(0) & 0(0) \\
\\
&\multirowcell{4}{$K=200$}&DSGC-S &  \textbf{0.10(0.447)} & 0(0) & 0.10(0.447) & 0.00(0.000) & 0(0) & 0(0) \\
&&DSC-S &   0.00(0.000) & 0(0) & 0.00(0.000) & 0.00(0.000) & 0(0) & 0(0) \\
&&GC-S &  0.00(0.000) & 0(0) & 0.00(0.000) & 0.00(0.000) & 0(0) & 0(0) \\
&&AC-S &  \textbf{7.40(5.651)} & 0(0) & 5.05(3.649) & 2.35(2.183) & 0(0) & 0(0) \\
 [-1.8ex] \\
\hline \\[-1.8ex]
\multirowcell{14}{$\sigma=0.5$}&\multirowcell{4}{$K=10$}&DSGC-S &   0.00(0.000) & \textbf{0.00(0.000)} & 0.00(0.000) & 0.0(0.000) & 0.00(0.000) & 0(0) \\
&&DSC-S &   0.00(0.000) & \textbf{0.05(0.224)} & 0.00(0.000) & 0.0(0.000) & 0.05(0.224) & 0(0) \\
&&GC-S &   0.00(0.000) & 0.00(0.000) & 0.00(0.000) & 0.0(0.000) & 0.00(0.000) & 0(0) \\
&&AC-S &   0.00(0.000) & 0.00(0.000) & 0.00(0.000) & 0.0(0.000) & 0.00(0.000) & 0(0) \\
\\
&\multirowcell{4}{$K=100$}&DSGC-S &  \textbf{0.10(0.308)} & 0.00(0.000) & 0.10(0.308) & 0.0(0.000) & 0.00(0.000) & 0(0) \\
&&DSC-S &   0.00(0.000) & 0.00(0.000) & 0.00(0.000) & 0.0(0.000) & 0.00(0.000) & 0(0) \\
&&GC-S &   0.00(0.000) & 0.00(0.000) & 0.00(0.000) & 0.0(0.000) & 0.00(0.000) & 0(0) \\
&&AC-S &  \textbf{3.55(5.073)} & 0.00(0.000) & 2.65(3.703) & 0.9(1.553) & 0.00(0.000) & 0(0) \\
\\
&\multirowcell{4}{$K=200$}&DSGC-S & \textbf{0.25(0.910)} & 0.00(0.000) & 0.15(0.671) & 0.1(0.308) & 0.00(0.000) & 0(0) \\
&&DSC-S &   0.00(0.000) & 0.00(0.000) & 0.00(0.000) & 0.0(0.000) & 0.00(0.000) & 0(0) \\
&&GC-S &   0.00(0.000) & 0.00(0.000) & 0.00(0.000) & 0.0(0.000) & 0.00(0.000) & 0(0) \\
&&AC-S &  \textbf{7.10(6.069)} & 0.00(0.000) & 4.80(3.847) & 2.3(2.342) & 0.00(0.000) & 0(0) \\
\hline\hline \\[-1.8ex]
\multicolumn{9}{l}{Note: The respective standard errors are recorded in the parentheses. }
\end{tabular}
\end{table}
\end{landscape}
\begin{landscape}
\begin{table}[!htbp] \centering
  \caption{Variable selection performance for \textbf{Case 2.} when the penalty function is determined as AL type.}
  \label{reAL2}
\begin{tabular}{@{\extracolsep{5pt}} ccccccccc}
\\[-1.8ex]\hline
\hline \\[-1.8ex]
&Number of Machines&Method & FP& FN & FP-C & FP-NC & FN-C & FN-NC \\
\hline \\[-1.8ex]
\multirowcell{14}{$\sigma=0.2$}&\multirowcell{4}{$K=10$}&DSGC-AL & 0.00(0.000) & 0(0) & 0.00(0.000) & 0.0(0.000) & 0(0) & 0(0) \\
&&DSC-AL &    0.00(0.000) & 0(0) & 0.00(0.000) & 0.0(0.000) & 0(0) & 0(0) \\
&&GC-AL &0.00(0.000) & 0(0) & 0.00(0.000) & 0.0(0.000) & 0(0) & 0(0) \\
&&AC-AL &  0.00(0.000) & 0(0) & 0.00(0.000) & 0.0(0.000) & 0(0) & 0(0) \\
\\
&\multirowcell{4}{$K=100$}&DSGC-AL & 0.00(0.000) & 0(0) & 0.00(0.000) & 0.0(0.000) & 0(0) & 0(0) \\
&&DSC-AL &   0.00(0.000) & 0(0) & 0.00(0.000) & 0.0(0.000) & 0(0) & 0(0) \\
&&GC-AL &  0.00(0.000) & 0(0) & 0.00(0.000) & 0.0(0.000) & 0(0) & 0(0) \\
&&AC-AL & 0.00(0.000) & 0(0) & 0.00(0.000) & 0.0(0.000) & 0(0) & 0(0) \\
\\
&\multirowcell{4}{$K=200$}&DSGC-AL &\textbf{0.00(0.000)} & 0(0) & 0.00(0.000) & 0.0(0.000) & 0(0) & 0(0) \\
&&DSC-AL &   0.00(0.000) & 0(0) & 0.00(0.000) & 0.0(0.000) & 0(0) & 0(0) \\
&&GC-AL &  0.00(0.000) & 0(0) & 0.00(0.000) & 0.0(0.000) & 0(0) & 0(0) \\
&&AC-AL &  \textbf{8.05(1.959)} & 0(0) & 5.65(1.531) & 2.4(1.046) & 0(0) & 0(0) \\
[-1.8ex] \\
\hline \\[-1.8ex]
\multirowcell{14}{$\sigma=0.5$}&\multirowcell{4}{$K=10$}&DSGC-AL &  0.00(0.000) & 0(0) & 0.00(0.000) & 0.00(0.000) & 0(0) & 0(0) \\
&&DSC-AL &     0.00(0.000) & 0(0) & 0.00(0.000) & 0.00(0.000) & 0(0) & 0(0) \\
&&GC-AL &  0.00(0.000) & 0(0) & 0.00(0.000) & 0.00(0.000) & 0(0) & 0(0) \\
&&AC-AL &  0.00(0.000) & 0(0) & 0.00(0.000) & 0.00(0.000) & 0(0) & 0(0) \\
\\
&\multirowcell{4}{$K=100$}&DSGC-AL & \textbf{0.00(0.000)} & 0(0) & 0.00(0.000) & 0.00(0.000) & 0(0) & 0(0) \\
&&DSC-AL &    0.00(0.000) & 0(0) & 0.00(0.000) & 0.00(0.000) & 0(0) & 0(0) \\
&&GC-AL & 0.00(0.000) & 0(0) & 0.00(0.000) & 0.00(0.000) & 0(0) & 0(0) \\
&&AC-AL &  \textbf{4.95(6.509)} & 0(0) & 3.25(4.241) & 1.70(2.296) & 0(0) & 0(0) \\
\\
&\multirowcell{4}{$K=200$}&DSGC-AL &  \textbf{0.15(0.366)} & 0(0) & 0.10(0.308) & 0.05(0.224) & 0(0) & 0(0) \\
&&DSC-AL &   0.00(0.000) & 0(0) & 0.00(0.000) & 0.00(0.000) & 0(0) & 0(0) \\
&&GC-AL &  0.00(0.000) & 0(0) & 0.00(0.000) & 0.00(0.000) & 0(0) & 0(0) \\
&&AC-AL &  \textbf{10.70(2.557)} & 0(0) & 7.40(1.465) & 3.30(1.559) & 0(0) & 0(0) \\
\hline\hline \\[-1.8ex]
\multicolumn{9}{l}{Note: The respective standard errors are recorded in the parentheses. }
\end{tabular}
\end{table}
\end{landscape}
\begin{landscape}
\begin{table}[!htbp] \centering
  \caption{Variable selection performance for \textbf{Case 2.} when the penalty function is determined as SCAD type.}
  \label{reS2}
\begin{tabular}{@{\extracolsep{5pt}} ccccccccc}
\\[-1.8ex]\hline
\hline \\[-1.8ex]
&Number of Machines&Method & FP& FN & FP-C & FP-NC & FN-C & FN-NC \\
\hline \\[-1.8ex]
\multirowcell{14}{$\sigma=0.2$}&\multirowcell{4}{$K=10$}&DSGC-S & 0.00(0.000) & 0(0) & 0.00(0.000) & 0.00(0.000) & 0(0) & 0(0) \\
&&DSC-S & 0.00(0.000) & 0(0) & 0.00(0.000) & 0.00(0.000) & 0(0) & 0(0) \\
&&GC-S &  0.00(0.000) & 0(0) & 0.00(0.000) & 0.00(0.000) & 0(0) & 0(0) \\
&&AC-S &  0.00(0.000) & 0(0) & 0.00(0.000) & 0.00(0.000) & 0(0) & 0(0) \\
\\
&\multirowcell{4}{$K=100$}&DSGC-S & \textbf{0.05(0.224)} & 0(0) & 0.05(0.224) & 0.00(0.000) & 0(0) & 0(0) \\
&&DSC-S &   0.00(0.000) & 0(0) & 0.00(0.000) & 0.00(0.000) & 0(0) & 0(0) \\
&&GC-S &  0.00(0.000) & 0(0) & 0.00(0.000) & 0.00(0.000) & 0(0) & 0(0) \\
&&AC-S &  \textbf{2.30(3.988)} & 0(0) & 1.95(3.154) & 0.35(0.875) & 0(0) & 0(0) \\
\\
&\multirowcell{4}{$K=200$}&DSGC-S & \textbf{0.55(0.826)} & 0(0) & 0.40(0.754) & 0.15(0.489) & 0(0) & 0(0) \\
&&DSC-S &   0.00(0.000) & 0(0) & 0.00(0.000) & 0.00(0.000) & 0(0) & 0(0) \\
&&GC-S &  0.00(0.000) & 0(0) & 0.00(0.000) & 0.00(0.000) & 0(0) & 0(0) \\
&&AC-S & \textbf{7.65(4.891)} & 0(0) & 6.40(3.831) & 1.25(1.773) & 0(0) & 0(0) \\ [-1.8ex] \\
\hline \\[-1.8ex]
\multirowcell{14}{$\sigma=0.5$}&\multirowcell{4}{$K=10$}&DSGC-S & 0.00(0.000) & 0(0) & 0.00(0.000) & 0.00(0.000) & 0(0) & 0(0) \\
&&DSC-S &  0.00(0.000) & 0(0) & 0.00(0.000) & 0.00(0.000) & 0(0) & 0(0) \\
&&GC-S &   0.00(0.000) & 0(0) & 0.00(0.000) & 0.00(0.000) & 0(0) & 0(0) \\
&&AC-S &  0.00(0.000) & 0(0) & 0.00(0.000) & 0.00(0.000) & 0(0) & 0(0) \\
\\
&\multirowcell{4}{$K=100$}&DSGC-S & \textbf{0.00(0.000)} & 0(0) & 0.00(0.000) & 0.00(0.000) & 0(0) & 0(0) \\
&&DSC-S &  0.00(0.000) & 0(0) & 0.00(0.000) & 0.00(0.000) & 0(0) & 0(0) \\
&&GC-S &   0.00(0.000) & 0(0) & 0.00(0.000) & 0.00(0.000) & 0(0) & 0(0) \\
&&AC-S &  \textbf{1.30(2.494)} & 0(0) & 1.25(2.314) & 0.05(0.224) & 0(0) & 0(0) \\
\\
&\multirowcell{4}{$K=200$}&DSGC-S & \textbf{0.75(1.251)} & 0(0) & 0.60(1.046) & 0.15(0.489) & 0(0) & 0(0) \\
&&DSC-S &  0.00(0.000) & 0(0) & 0.00(0.000) & 0.00(0.000) & 0(0) & 0(0) \\
&&GC-S &  0.00(0.000) & 0(0) & 0.00(0.000) & 0.00(0.000) & 0(0) & 0(0) \\
&&AC-S &  \textbf{7.95(4.273)} & 0(0) & 6.95(3.486) & 1.00(1.556) & 0(0) & 0(0) \\
\hline\hline \\[-1.8ex]
\multicolumn{9}{l}{Note: The respective standard errors are recorded in the parentheses. }
\end{tabular}
\end{table}
\end{landscape}
\subsection{Real Data}
The medical insurance(MI) dataset for major chest surgeries contains information about patients who underwent major chest surgery at a large tertiary hospital in Sichuan Province, China, from January 2020 to December 2020. MI dataset is collected by Comprehensive Hospital of a Class-3 Grade-A Hospital in Chengdu, China, and available with reasonable request.  The patients were grouped according to the Chinese Diagnosis Related Groups (CHS-DRG) standard for diagnosis and treatment under the National Medical Insurance scheme. Our objective is to predict the medical insurance reimbursement ratio(MIRR) for patients. The MIRR is defined as the ratio of the reimbursement amount(RA) to the total hospitalization costs(THC). After eliminating any values that are not available for RA and THC, the total sample size of MI dataset is $n=37342$. We define hospitalization costs as the explanatory variable, which is composed of $p=15$ components: income from western medicine (IWM), income from diagnosis (ID), income from examination (IE), income from radiology (IR), income from treatment (IT), income from surgery (IS), income from laboratory tests (ILT), income from nursing (IN), income from bed occupancy (IBO), income from medical supplies (IMS), income from Chinese patent medicine (ICP), income from Chinese herbal medicine (ICHM), income from nutrition and meals (INM), income from blood products (IBP), and income from oxygen supplies (IOS). We exclude variables that are not related to treatment costs, such as patient age, medical insurance type, admitting department, and discharge department. Additionally, we add $q=20$ non-compositional variables $\cude{V}=(V_1,\dots,V_{20})\in\mathbbm{R}^{n\times q}$ that are not related to the MIRR, where $\cude{V}$ is generated from a multivariate $t$ distribution $\cude{T}_3(\cude{0}, \cude{\Sigma})$ with $3$ degrees of freedom and $\cude{\Sigma}=(\sigma^{(i\neq j)})$ with $\sigma\in\{0.2, 0.5\}$.\par
In this subsection, because the true parameter is unknown and the computing procedure is carried on a single machine, we first evaluate the prediction accuracy of our proposed methods DSCDMM and DSGCDMM, and compare them with GCDMM and ACDMM proposed in Section \ref{NE}. The penalty function is specified as adaptive LASSO and SCAD. We artificially split the PTS data set into $K=\{10, 20, 50\}$ subdata sets, and the other simulation settings are same as those in Section \ref{NE}. To obtain stable and reliable prediction results, we execute the 5-folds cross-validation(CV) to evaluate the performance. More specifically, the CV prediction error is measured by the $\ell_q$ losses, i.e.,  $$\text{CVL}_q=\frac{1}{5}\sum_{l=1}^5\frac{\|\cude{y}_{v_l}-\cude{\Pi}_{v_l}\widehat{\cude{\zeta}}_{l}\|_q}{n_{v_l}}$$ with $q=1, 2, \infty$, where $\{\cude{y}_{v_l}, \cude{\Pi}_{v_l}\}$ is the validation dataset for $l^{th}$ fold and $n_{v_l}$ is its sample size, $\widehat{\cude{\zeta}}_{l}$ is obtained by fitting the training datasets for $l^{th}$ fold. We also calculate the average values of FN-NC and FP-NC(as defined in Section \ref{NE}), which are relabeled as CVFN-NC and CVFP-NC, respectively, by using the training dataset for each fold in the CV procedure. This allows us to validate the accuracy of variable selection. Tables \ref{rl1}---\ref{rl2} present the average values of three metrics, namely, $\text{CVL}_q$, CVFN-NC, and CVFP-NC, based on 100 simulations for  $\sigma=0.2$ and $\sigma=0.5$, respectively. By examining the tables for different  values of $\sigma$ and $K$, we observe that the values of $\text{CVL}_q$ for DSGCDMM and DSCDMM, with both adaptive LASSO and SCAD penalties, are extremely close to those of the global method GCDMM and the average-based method ACDMM. Additionally, we note that the values of CVFP-NC for ACDMM become increasingly larger than those of other two distributed methods DSGCDMM and DSCDMM as the number of machines increases. This suggests that average-based methods with only one round of communication exhibit lower statistical efficiency when the number of machines is large. Besides, although the values of DSGCDMM also increase as the number of machines increases, DSGCDMM has more stable and much smaller CVFP-NC compared to ACDMM. We can draw a conclusion that our proposed DSGCDMM and DSCDMM have good prediction performance and demonstrates its practicality and effectiveness for handling real-world data. It is noted that DSGCDMM and DSCDMM have different applicable scenarios, namely centralized and decentralized frameworks. Users can choose the appropriate method based on the specific scenario.
\begin{table}[!htbp] \centering 
  \caption{The simulation results for the MI data over 100 replications, where $\sigma=0.2$.} 
  \label{rl1} 
\begin{tabular}{@{\extracolsep{5pt}} ccccccc} 
\\[-1.8ex]\hline 
\hline \\[-1.8ex] 
&Method&$\text{CVL}_1$ & $\text{CVL}_2$ & $\text{CVL}_\infty$ &  CVFP-NC &  CVFN-NC \\ 
\hline \\[-1.8ex] 
\multirowcell{8}{$K=10$}&DSGC-AL&$0.28076$ & $0.00449$ & $0.00016$ & $0$ & $0$ \\ 
&DSC-AL&$0.30293$ & $0.00480$ & $0.00016$ & $0$ & $0$ \\ 
&GC-AL&$0.27161$ & $0.00433$ & $0.00019$ & $0$ & $0$ \\ 
&AC-AL&$0.27321$ & $0.00436$ & $0.00016$ & $0$ & $0$ \\
\cline{2-7}
&DSGC-S&$0.26604$ & $0.00427$ & $0.00019$ & $0.43$ & $0$ \\ 
&DSC-S&$0.31042$ & $0.00489$ & $0.00020$ & $0$ & $0$ \\ 
&GC-S&$0.26385$ & $0.00424$ & $0.00028$ & $0$ & $0$ \\ 
&AC-S&$0.26323$ & $0.00423$ & $0.00020$ & $2.87$ & $0$ \\ 
\hline
\multirowcell{8}{$K=20$}&DSGC-AL&$0.28190$ & $0.00450$ & $0.00016$ & $0$ & $0$ \\ 
&DSC-AL&$0.30178$ & $0.00479$ & $0.00016$ & $0$ & $0$ \\ 
&GC-AL&$0.27164$ & $0.00433$ & $0.00019$ & $0$ & $0$ \\ 
&AC-AL&$0.26560$ & $0.00426$ & $0.00020$ & $0$ & $0$ \\ 
\cline{2-7}
&DSGC-S&$0.26590$ & $0.00425$ & $0.00019$ & $1.32$ & $0$ \\ 
&DSC-S&$0.30535$ & $0.00486$ & $0.00016$ & $0$ & $0$ \\ 
&GC-S&$0.26381$ & $0.00424$ & $0.00028$ & $0$ & $0$ \\ 
&AC-S&$0.26489$ & $0.00432$ & $0.00048$ & $13.21$ & $0$ \\ 
\hline
\multirowcell{8}{$K=50$}&DSGC-AL&$0.27966$ & $0.00447$ & $0.00016$ & $0$ & $0$ \\ 
&DSC-AL&$0.15684$ & $0.00255$ & $0.00010$ & $0$ & $0$ \\ 
&GC-AL&$0.27162$ & $0.00433$ & $0.00019$ & $0$ & $0$ \\ 
&AC-AL&$0.25711$ & $0.00414$ & $0.00019$ & $10.05$ & $0$ \\
\cline{2-7}
&DSGC-S&$0.27341$ & $0.00437$ & $0.00025$ & $3.99$ & $0$ \\ 
&DSC-S&$0.30150$ & $0.00478$ & $0.00016$ & $0$ & $0$ \\ 
&GC-S&$0.26385$ & $0.00424$ & $0.00028$ & $0$ & $0$ \\ 
&AC-S&$0.25260$ & $0.00409$ & $0.00025$ & $19.32$ & $0$ \\ 
\hline\hline \\[-1.8ex] 
\end{tabular} 
\end{table} 
\begin{table}[!htbp] \centering 
  \caption{The simulation results for the MI data over 100 replications, where $\sigma=0.5$.} 
  \label{rl2} 
\begin{tabular}{@{\extracolsep{5pt}} ccccccc} 
\\[-1.8ex]\hline 
\hline \\[-1.8ex] 
&Method&$\text{CVL}_1$ & $\text{CVL}_2$ & $\text{CVL}_\infty$ &  CVFP-NC &  CVFN-NC \\ 
\hline \\[-1.8ex] 
\multirowcell{8}{$K=10$}&DSGC-AL&$0.28083$ & $0.00449$ & $0.00016$ & $0$ & $0$ \\ 
&DSC-AL&$0.30314$ & $0.00480$ & $0.00016$ & $0$ & $0$ \\ 
&GC-AL&$0.27151$ & $0.00433$ & $0.00019$ & $0$ & $0$ \\ 
&AC-AL&$0.27322$ & $0.00436$ & $0.00016$ & $0$ & $0$ \\ 
\cline{2-7}
&DSGC-S&$0.26603$ & $0.00427$ & $0.00019$ & $0.33$ & $0$ \\ 
&DSC-S&$0.31037$ & $0.00489$ & $0.00020$ & $0$ & $0$ \\ 
&GC-S&$0.26381$ & $0.00424$ & $0.00028$ & $0$ & $0$ \\ 
&AC-S&$0.26259$ & $0.00422$ & $0.00021$ & $3.13$ & $0$ \\ 
\hline
\multirowcell{8}{$K=20$}&DSGC-AL&$0.28192$ & $0.00450$ & $0.00016$ & $0$ & $0$ \\ 
&DSC-AL&$0.30180$ & $0.00479$ & $0.00016$ & $0$ & $0$ \\ 
&GC-AL&$0.27158$ & $0.00433$ & $0.00019$ & $0$ & $0$ \\ 
&AC-AL&$0.26561$ & $0.00426$ & $0.00020$ & $0.04$ & $0$ \\ 
\cline{2-7}
&DSGC-S&$0.26901$ & $0.00430$ & $0.00020$ & $1.39$ & $0$ \\ 
&DSC-S&$0.30203$ & $0.00482$ & $0.00016$ & $0$ & $0$ \\ 
&GC-S&$0.26353$ & $0.00424$ & $0.00028$ & $0$ & $0$ \\ 
&AC-S&$0.26464$ & $0.00432$ & $0.00048$ & $12.79$ & $0$ \\ 
\hline
\multirowcell{8}{$K=50$}&DSGC-AL&$0.27977$ & $0.00447$ & $0.00016$ & $0$ & $0$ \\ 
&DSC-AL&$0.15698$ & $0.00256$ & $0.00010$ & $0$ & $0$ \\ 
&GC-AL&$0.27160$ & $0.00433$ & $0.00019$ & $0$ & $0$ \\ 
&AC-AL&$0.25763$ & $0.00415$ & $0.00019$ & $11.85$ & $0$ \\
\cline{2-7}
&DSGC-S&$0.27306$ & $0.00437$ & $0.00026$ & $3.75$ & $0$ \\ 
&DSC-S&$0.30177$ & $0.00479$ & $0.00016$ & $0$ & $0$ \\ 
&GC-S&$0.26385$ & $0.00424$ & $0.00028$ & $0$ & $0$ \\ 
&AC-S&$0.25268$ & $0.00409$ & $0.00025$ & $19$ & $0$ \\ 
\hline\hline \\[-1.8ex] 
\end{tabular} 
\end{table}  
\section{Concluding Remarks}\label{CON}
In this paper, we propose variable selection procedures for the distributed massive linear regression with compositional covariates, and we consider two distributed frameworks with centralized and decentralized topologies. In the centralized manner, by combining ADMM(\citet{boyd2011distributed}) and CDMM(\citet{lin2014variable}) algorithms, we solve the distributed optimization problem with two types of constraints that are related to the global model parameter. We also present the  associated statistical efficient algorithm, named DSCDMM. To address the drawbacks of the centralized distributed system, we further propose a novel decentralized approach, namely DSGCDMM, which is motivated from the GADMM algorithm(\cite{elgabli2020gadmm}). The DSGCDMM algorithm can solve the distributed constrained optimization problem that is only related to the local machine parameters. Simulation studies and a real data example demonstrate that our proposed DSCDMM and DSGCDMM algorithms can exhibit their statistical efficiency in variable selection procedures under centralized and decentralized manners. In summary, both of these two algorithms have their own applicable scenarios in practice.\par
However, this work has several limitations. First, the DSCDMM and DSGCDMM algorithms perform less computational efficiency according to the running times. Second, in the high-dimensional or ultra high-dimensional settings, our proposed approaches may not work well, particularly when there is a high computational burden due to multiple rounds of communication and the relatively low computational efficiency of coordinate-wise descent algorithms. Especially, for problems with a large number of variables, coordinate-wise descent algorithms can be very inefficient. This is because they require evaluating the objective function and its gradient at each iteration, which can be very computationally expensive for large problems(\cite{wright2015coordinate}). In addition,  the heterogeneity for the distributed datasets from machine to machine and privacy-preserving issues in both distributed learning and federated learning(\cite{konevcny2016federated,kairouz2021advances}) are not considered in our framework(\cite{duan2022heterogeneity,zhu2021least}). Therefore, developing a more fast and communication-efficient methodology  together with efficient statistical estimation and computation under the above issues is our next objective for next work. Moreover, there are many interesting and valuable research opportunities in the field of decentralized distributed machine learning that are worth pursuing in the future.
\acks{Ma’s research was fully supported by the National Natural Science Foundation of China
(Grant No.12101439) and the Natural Science Foundation of Jiangsu Province (Grant No.BK20200854). Huang’s research was fully supported by the Sichuan Natural Science Foundation (Grant No.2022NSFSC1850), the New Interdisciplinary Training Fund(Grant No.2682023JX004), and partially supported by National Natural Science Foundation of China(Grant No.72033002).}


\newpage

\appendix
\section{}
\label{app1}


\noindent
{\bf Proof of Lemma \ref{lem1}} 
{It is noted that the  augmented Lagrangian function $\mathscr{L}_{\rho}$ in \eqref{lm} is sub-differentiable. Under the proofs of \cite{elgabli2020gadmm}, for the minimizers $\cude{\zeta}^{l+1}_k$'s with $k\in \mathscr{N}_h$ of \eqref{nu2} ans \eqref{nu1}, the expressions \eqref{ph1} and \eqref{ph2} can be rewritten as 
\begin{equation}\label{ph1n}
      \cude{0}\in \partial \mathbb{Q}_k(\cude{\zeta}_k^{l+1}) + \mu_{k}^{l+1}\cude{C}+\cude{\gamma}_k^{l+1}+\cude{s}_k^{l+1},~k=1.
\end{equation}
\begin{equation}\label{ph2n}
    \cude{0}\in \partial \mathbb{Q}_k(\cude{\zeta}_k^{l+1}) + \mu_{k}^{l+1}\cude{C}-\cude{\gamma}_{k-1}^{l+1}+\cude{\gamma}_k^{l+1}+\cude{s}_k^{l+1},~k\in \mathscr{N}_h\backslash\{1\}.
\end{equation}
Following the technical proof of \textbf{Lemma 1} in \cite{elgabli2020gadmm}, we also denote $\cude{\gamma}_0^{l+1}=\cude{\gamma}_K^{l+1}=\cude{0}$ for simplicity. It is observed that $\cude{\zeta}_{k\in\mathscr{N}_h}^{l+1}$ and $\cude{\zeta}_{k\in\mathscr{N}_t}^{l+1}$ can be regarded as the minimizers of the following two convex objective functions $\mathbb{H}_k(\cude{\zeta}_k)$ and $\mathbb{T}_k(\cude{\zeta}_k)$, respectively. They are separately defined as
\begin{equation}\label{hef}
\mathbb{H}_k(\cude{\zeta}_k):=\mathbb{Q}_k(\cude{\zeta}_k)+\left\langle \mu_{k}^{l+1}\cude{C}-\cude{\gamma}_{k-1}^{l+1}+\cude{\gamma}_k^{l+1}+\cude{s}_k^{l+1}, \cude{\zeta}_k\right\rangle,
\end{equation}
and 
\begin{equation}\label{taf}
\mathbb{T}_k(\cude{\zeta}_k):=\mathbb{Q}_k(\cude{\zeta}_k)+\left\langle \mu_{k}^{l+1}\cude{C}-\cude{\gamma}_{k-1}^{l+1}+\cude{\gamma}_k^{l+1}, \cude{\zeta}_k\right\rangle.
\end{equation}
Thus, for the optimal solution $\cude{\zeta}^\star$ of the problem in \eqref{optpl}, we can obtain 
\begin{align}
   \mathbb{H}_k(\cude{\zeta}_k^{l+1})&\leq \mathbb{H}_k(\cude{\zeta}_k^{\star}),  \\
    \mathbb{T}_k(\cude{\zeta}_k^{l+1})&\leq \mathbb{T}_k(\cude{\zeta}_k^{\star}). 
\end{align}
It follows that
\begin{equation}\label{sumlht}
\begin{split}
&\sum_{k\in\mathscr{N}_h}\mathbb{H}_k(\cude{\zeta}_k^{l+1})+\sum_{k\in\mathscr{N}_t}\mathbb{T}_k(\cude{\zeta}_k^{l+1})\leq \sum_{k\in\mathscr{N}_h}\mathbb{H}_k(\cude{\zeta}_k^{\star})+\sum_{k\in\mathscr{N}_t}\mathbb{T}_k(\cude{\zeta}_k^{\star}),
\end{split}
\end{equation}
After rearranging the terms and substituting the expression from \eqref{hef} and \eqref{taf}, \eqref{sumlht} can be re-expressed as 
\begin{equation}
    \begin{split}
        &\sum_{k=1}^K\mathbb{Q}_k(\cude{\zeta}_k^{l+1})-\sum_{k=1}^K\mathbb{Q}_k(\cude{\zeta}_k^{\star})\\
        &\leq \sum_{k=1}^K\left\langle \mu_k^{l+1}\cude{C},\cude{\zeta}_k^{\star}\right\rangle- \sum_{k=1}^K\left\langle \mu_k^{l+1}\cude{C},\cude{\zeta}_k^{l+1}\right\rangle+\sum_{k\in\mathscr{N}_t}\left\langle-\cude{\gamma}_{k-1}^{l+1}+\cude{\gamma}_k^{l+1}, \cude{\zeta}^\star\right\rangle+\sum_{k\in\mathscr{N}_h}\left\langle-\cude{\gamma}_{k-1}^{l+1}+\cude{\gamma}_k^{l+1}, \cude{\zeta}^\star\right\rangle\\
        &-\sum_{k\in\mathscr{N}_t}\left\langle-\cude{\gamma}_{k-1}^{l+1}+\cude{\gamma}_k^{l+1}, \cude{\zeta}_k^{l+1}\right\rangle-\sum_{k\in\mathscr{N}_h}\left\langle-\cude{\gamma}_{k-1}^{l+1}+\cude{\gamma}_k^{l+1}, \cude{\zeta}_k^{l+1}\right\rangle+\sum_{k\in\mathscr{N}_h}\left\langle \cude{s}_k^{l+1}, \cude{\zeta}^\star-\cude{\zeta}^{l+1}_k\right\rangle.
    \end{split}
\end{equation}
Through the proof of \textbf{Lemma 1} in \cite{elgabli2020gadmm} and the fact $\cude{C}^\top\cude{\zeta}^\star_k=0$, we have
$$\sum_{k\in\mathscr{N}_t}\left\langle-\cude{\gamma}_{k-1}^{l+1}+\cude{\gamma}_k^{l+1}, \cude{\zeta}^\star\right\rangle+\sum_{k\in\mathscr{N}_h}\left\langle-\cude{\gamma}_{k-1}^{l+1}+\cude{\gamma}_k^{l+1}, \cude{\zeta}^\star\right\rangle=0,$$
$$\sum_{k\in\mathscr{N}_t}\left\langle-\cude{\gamma}_{k-1}^{l+1}+\cude{\gamma}_k^{l+1}, \cude{\zeta}_k^{l+1}\right\rangle+\sum_{k\in\mathscr{N}_h}\left\langle-\cude{\gamma}_{k-1}^{l+1}+\cude{\gamma}_k^{l+1}, \cude{\zeta}_k^{l+1}\right\rangle=\sum_{k=1}^{K-1}\left\langle\cude{\gamma}_k^{l+1},\cude{r}_k^{l+1}\right\rangle.$$
Therefore, we can obtain
\begin{equation}
\begin{split}
     \sum_{k=1}^K\mathbb{Q}_k(\cude{\zeta}_k^{l+1})-\sum_{k=1}^K\mathbb{Q}_k(\cude{\zeta}_k^{\star})&\leq - \sum_{k=1}^K\left\langle \mu_k^{l+1}\cude{C},\cude{\zeta}_k^{l+1}\right\rangle-\sum_{k=1}^{K-1}\left\langle\cude{\gamma}_k^{l+1},\cude{r}_k^{l+1}\right\rangle+\sum_{k\in\mathscr{N}_h}\left\langle \cude{s}_k^{l+1}, \cude{\zeta}^\star-\cude{\zeta}^{l+1}_k\right\rangle\\
    & =- \sum_{k=1}^K\mu_k^{l+1}g_k^{l+1}-\sum_{k=1}^{K-1}\left\langle\cude{\gamma}_k^{l+1},\cude{r}_k^{l+1}\right\rangle+\sum_{k\in\mathscr{N}_h}\left\langle \cude{s}_k^{l+1}, \cude{\zeta}^\star-\cude{\zeta}^{l+1}_k\right\rangle.
\end{split}
\end{equation}
 According to the proof steps of \cite{boyd2011distributed,giesen2016distributed,elgabli2020gadmm}, for a saddle point $(\cude{\zeta}^\star,\{\mu_k^\star\}_{k=1}^K, \{\cude{\gamma}_k^\star\}_{k=1}^K)$ of $\mathscr{L}_0\left(\{\cude{\zeta}_k\}^K_{k=1},\{\mu_k\}_{k=1}^K,\{\cude{\gamma}_k\}_{k=1}^K\right)$, the following inequality holds:
 \begin{equation}\label{inL}
  \mathscr{L}_0\left(\cude{\zeta}^\star,\{\mu_k^\star\}_{k=1}^K,\{\cude{\gamma}_k^\star\}_{k=1}^K\right)\leq  \mathscr{L}_0\left(\{\cude{\zeta}_k^{l+1}\}^K_{k=1},\{\mu_k^\star\}_{k=1}^K,\{\cude{\gamma}_k^\star\}_{k=1}^K\right). 
 \end{equation}
 Using the expression for the augmented Lagrangian as given in equation \eqref{lm} to substitute on both sides of equation \eqref{inL}, it follows that
  \begin{equation}
  \begin{split}
\sum_{k=1}^K\left(\mathbb{Q}_k(\cude{\zeta}_k^{l+1})-\mathbb{Q}_k(\cude{\zeta}_k^{\star})\right)&\geq -\sum_{k=1}^K\left\langle\mu_k^\star\cude{C}^\top, \cude{\zeta}_k^{l+1}\right\rangle-\sum_{k=1}^{K-1}\left\langle\cude{\gamma}_{k}^\star,\cude{r}_k^{l+1}\right\rangle\\
&=-\sum_{k=1}^K\mu_k^\star g_k^{l+1}-\sum_{k=1}^{K-1}\left\langle\cude{\gamma}_{k}^\star,\cude{r}_k^{l+1}\right\rangle.
  \end{split}
    \end{equation}
    This completes the proof of the lemma.}
\hfill\BlackBox
\section{}
\label{app2}
\noindent
{\bf Proof of Theorem \ref{thm1}}.
    {Our procedures of proof likewise follow the steps of \cite{boyd2011distributed,elgabli2020gadmm}. From \textbf{Lemma 1}, we can obtain
    \begin{equation}\label{thl1}
         -\sum_{k=1}^K\mu_k^\star g_k^{l+1}-\sum_{k=1}^{K-1}\left\langle\cude{\gamma}_{k}^\star,\cude{r}_k^{l+1}\right\rangle\leq  -\sum_{k=1}^K\mu_k^{l+1}g_k^{l+1}-\sum_{k=1}^{K-1}\left\langle\cude{\gamma}_k^{l+1},\cude{r}_k^{l+1}\right\rangle+\sum_{k\in\mathscr{N}_h}\left\langle\cude{s}^{l+1}_k, \cude{\zeta}_k^\star-\cude{\zeta}_k^{l+1}\right\rangle.
    \end{equation} 
    After rearranging \eqref{thl1} and multiplying both sides of \eqref{thl1} by $2$, we have
    \begin{equation}
    \begin{split}
         &2\sum_{k=1}^K(\mu_k^{l+1}-\mu_k^\star)g_k^{l+1}+2\sum_{k=1}^{K-1}\left\langle\cude{\gamma}_k^{l+1}-\cude{\gamma}_k^\star,\cude{r}_k^{l+1}\right\rangle+2\sum_{k\in\mathscr{N}_h}\left\langle\cude{s}_k^{l+1},\cude{\zeta}_k^{l+1}-\cude{\zeta}_k^\star\right\rangle\leq 0.
    \end{split}
    \end{equation}
    The update of dual variable $\mu_k^{l+1}$ in \eqref{nu5} can be rewritten as $\mu_k^{l+1}=\mu_k^l+\rho g_k^{l+1}$. Then $g_k^{l+1}=(\mu_k^{l+1}-\mu_k^l)/\rho$. Besides, 
    $$\mu_k^{l+1}-\mu_k^l=(\mu_k^{l+1}-\mu_k^\star)-(\mu_k^{l}-\mu_k^\star).$$
    Direct calculation yields
    \begin{equation}\label{THA1}
    \begin{split}
    2\sum_{k=1}^K(\mu_k^{l+1}-\mu_k^\star)g_k^{l+1}&=2\sum_{k=1}^K(\mu_k^l+\rho g_k^{l+1}-\mu_k^\star)g_k^{l+1}  \\
    &=2\sum_{k=1}^K(\mu_k^l-\mu_k^\star)g_k^{l+1}+\rho\sum_{k=1}^K(g_k^{l+1})^2+\rho\sum_{k=1}^K(g_k^{l+1})^2\\
    &=\frac{2}{\rho}\sum_{k=1}^K(\mu_k^l-\mu_k^\star)(\mu_k^{l+1}-\mu_k^l)+\frac{1}{\rho}\sum_{k=1}^K(\mu_k^{l+1}-\mu_k^l)^2+\rho\sum_{k=1}^K(g_k^{l+1})^2\\
    &=\frac{1}{\rho}\sum_{k=1}^K(\mu_k^{l+1}-\mu_k^\star)^2-\frac{1}{\rho}\sum_{k=1}^K(\mu_k^{l}-\mu_k^\star)^2+\rho\sum_{k=1}^K(g^{l+1}_k)^2\\
    &:=A_1.
    \end{split}
    \end{equation}
    Applying the similar proof technique as in \eqref{THA1}, it holds that
    \begin{equation}
        \begin{split}
            2\sum_{k=1}^{K-1}\left\langle\cude{\gamma}_k^{l+1}-\cude{\gamma}_k^\star,\cude{r}_k^{l+1}\right\rangle&=\frac{1}{\rho}\sum_{k=1}^{K-1}\left\|\cude{\gamma}_k^{l+1}-\cude{\gamma}_k^\star\right\|^2-\frac{1}{\rho}\sum_{k=1}^{K-1}\left\|\cude{\gamma}_k^{l+1}-\cude{\gamma}_k^\star\right\|^2+\rho\sum_{k=1}^{K-1}\left\|\cude{r}_k^{l+1}\right\|^2\\
            &:=A_2.
        \end{split}
    \end{equation}
    Next, following the proof of \textbf{Theorem 2} in \cite{elgabli2020gadmm}, we have
    \begin{equation}
        \begin{split}
&~2\sum_{k\in\mathscr{N}_h}\left\langle\cude{s}_k^{l+1},\cude{\zeta}_k^{l+1}-\cude{\zeta}_k^\star\right\rangle\\
&=\sum_{n\in{\mathscr{N}}_{h}\setminus\{1\}}\left(\begin{array}{l}-2\rho\left\langle\cude{\zeta}_{k-1}^{l+1}-\cude{\zeta}_{k-1}^{l},\cude{r}_{k-1}^{l+1}\right\rangle+\rho\left\|\boldsymbol{\zeta}_{k-1}^{l+1}-\boldsymbol{\zeta}_{k-1}^{l}\right\|^{2}+\rho\left\|\cude{\zeta}^{l+1}_{k-1}-\cude{\zeta}_{k-1}^l\right\|^2\\
~~~~~~~~+\rho\left\|\cude{\zeta}_{k-1}^{l+1}-\cude{\zeta}^\star\right\|^2-\rho\left\|\cude{\zeta}^l_{k-1}-\cude{\zeta}^\star\right\|^2
\end{array}\right)\\
&~~~+\sum\limits_{n\in\mathscr{N}_h}\left(2\rho\left\langle\cude{\zeta}_{k+1}^{l+1}-\cude{\zeta}_{k+1}^{l},\cude{r}_{k}^{l+1}\right\rangle+\rho\left\|\cude{\zeta}_{k+1}^{l+1}-\cude{\zeta}_{k+1}^l\right\|^2+\rho\left\|\cude{\zeta}_{k+1}^{l+1}-\cude{\zeta}^\star\right\|^2-\rho\left\|\cude{\zeta}^l_{k+1}-\cude{\zeta}^\star\right\|^2\right)\\
&:=A_3.
        \end{split}
    \end{equation}
Now we define a Lyapunov function $V^l$ for our algorithm as
\begin{equation}\label{Lya}
    V^l=\frac{1}{\rho}\sum_{k=1}^K(\mu_k^l-\mu_k^\star)^2+\frac{1}{\rho}\sum_{k=1}^{K-1}\left\|\cude{\gamma}_k^l-\cude{\gamma}_k^\star\right\|^2+\rho\sum_{k\in\mathscr{N}_h\setminus\{1\}}\left\|\cude{\zeta}_{k-1}^l-\cude{\zeta}^\star\right\|^2+\rho\sum_{k\in\mathscr{N}_h}\left\|\cude{\zeta}_{k+1}^l-\cude{\zeta}^\star\right\|^2.
\end{equation}
After rearranging the terms in inequality $A_1+A_2+A_3\geq 0$ and through \eqref{Lya}, the following inequality holds
\begin{equation}\label{diLy}
\begin{split}
    V^l-V^{l+1}&\geq \rho\sum_{k=1}^K(g_k^{l+1})^2+\rho\sum_{k=1}^{K-1}\left\|\cude{r}_k^{l+1}\right\|^2+\Lambda^l,
    \end{split}
\end{equation}
where 
\begin{equation}\label{VL}
\begin{split}
\Lambda^l&=\left[\rho\sum_{k\in\mathscr{N}_h\setminus\{1\}}\left\|\cude{\zeta}^{l+1}_{k-1}-\cude{\zeta}_{k-1}^l\right\|^2+\rho\sum_{k\in\mathscr{N}_h}\left\|\cude{\zeta}_{k+1}^{l+1}-\cude{\zeta}_{k+1}^l\right\|^2\right]\\
    &~~~~+\left[-2\rho\sum_{n\in\mathscr{N}_{h}\setminus\{1\}}\left\langle\cude{\zeta}_{k-1}^{l+1}-\cude{\zeta}_{k-1}^{l},\cude{r}_{k-1}^{l+1}\right\rangle+2\rho\sum_{k\in\mathscr{N}_{h}}\left\langle\cude{\zeta}_{k+1}^{l+1}-\boldsymbol{\zeta}_{k+1}^{l},\cude{r}_{k}^{l+1}\right\rangle\right]\\
    &:=\Lambda_1^l+\Lambda_2^l.
    \end{split}
\end{equation}
This indicates that $V_l$ decreases monotonically at each iteration as long as $\Lambda_1^l$,$\Lambda_2^l$ in \eqref{VL} are positive. It is obvious that $\Lambda_1>0$. Applying the proof results directly of \textbf{Theorem 2} in \cite{elgabli2020gadmm}, we obtain $\Lambda_2>0$ holds. Then,
\begin{equation}\label{diLyn}
\begin{split}
    V^l-V^{l+1}&\geq \rho\sum_{k=1}^K(g_k^{l+1})^2+\rho\sum_{k=1}^{K-1}\left\|\cude{r}_k^{l+1}\right\|^2+\Lambda_1^l.
    \end{split}
\end{equation}
For $l=0,\ldots,L$, $V^L\leq V^0$ due to the monotonicity decreasing of $V^l$. Sum both sides of \eqref{diLyn} over $l$, we have
\begin{equation}\label{serLy}
\sum_{l=0}^L\left[\rho\sum_{k=1}^K(g_k^{l+1})^2+\rho\sum_{k=1}^{K-1}\left\|\cude{r}_k^{l+1}\right\|^2+\Lambda_1^l\right]\leq V^0-V^L\leq V^0.
\end{equation}
There is a fact that a series of positive terms that is bounded converges. We infer that the series in the left side of \eqref{serLy} converges as $L\to\infty$. Hence, $g_k^{l+1}\to 0$ for all $k=1,\ldots,K$ and $r_k^{l+1}\to \cude{0}$ for all $k=1,\ldots, K-1$, as $l\to \infty$. This completes the proof of propositions \textit{1.} and \textit{2.} in \textbf{Theorem \ref{thm1}}. Furthermore, the fact that $\left\|\cude{\zeta}^{l+1}_{k-1}-\cude{\zeta}_{k-1}^l\right\|\to 0$ and $\left\|\cude{\zeta}_{k+1}^{l+1}-\cude{\zeta}_{k+1}^l\right\|\to 0$ as $l\to\infty$ results in $\cude{s}_k^l\to\cude{0}$. This completes the proof of proposition \textit{3.} in \textbf{Theorem \ref{thm1}}. By the \textbf{Lemma \ref{lem1}}, the propositions \textit{1.}, \textit{2.}, \textit{3.} in \textbf{Theorem \ref{thm1}} and Squeeze theorem in \cite{sohrab2003basic}, it holds that, as $l\to\infty$,  $$\sum_{k=1}^K\left(\mathbb{Q}_k(\cude{\zeta}_k^{l})-\mathbb{Q}_k(\cude{\zeta}_k^{\star})\right)\rightarrow 0.$$
This completes the proof of the theorem.}
\hfill\BlackBox

\vskip 0.2in
\bibliography{sample}

\begin{thebibliography}{60}
\providecommand{\natexlab}[1]{#1}
\providecommand{\url}[1]{\texttt{#1}}
\expandafter\ifx\csname urlstyle\endcsname\relax
  \providecommand{\doi}[1]{doi: #1}\else
  \providecommand{\doi}{doi: \begingroup \urlstyle{rm}\Url}\fi

\bibitem[Aitchison(1982)]{aitchison1982statistical}
John Aitchison.
\newblock The statistical analysis of compositional data.
\newblock \emph{Journal of the Royal Statistical Society: Series B (Satistical
  Methodology)}, 44\penalty0 (2):\penalty0 139--160, 1982.

\bibitem[Aitchison and Bacon-Shone(1984)]{aitchison1984log}
John Aitchison and John Bacon-Shone.
\newblock Log contrast models for experiments with mixtures.
\newblock \emph{Biometrika}, 71\penalty0 (2):\penalty0 323--330, 1984.

\bibitem[Alenazi(2021)]{alenazi2021review}
Abdulaziz Alenazi.
\newblock A review of compositional data analysis and recent advances.
\newblock \emph{Communications in Statistics-Theory and Methods}, pages 1--33,
  2021.

\bibitem[Atallah et~al.(2022)Atallah, Rahnavard, and Sun]{atallah2022codgrad}
Elie Atallah, Nazanin Rahnavard, and Qiyu Sun.
\newblock {CoDGraD}: A code-based distributed gradient descent scheme for
  decentralized convex optimization.
\newblock \emph{arXiv preprint arXiv:2204.06344}, 2022.

\bibitem[Boyd et~al.(2011)Boyd, Parikh, Chu, Peleato, Eckstein,
  et~al.]{boyd2011distributed}
Stephen Boyd, Neal Parikh, Eric Chu, Borja Peleato, Jonathan Eckstein, et~al.
\newblock Distributed optimization and statistical learning via the alternating
  direction method of multipliers.
\newblock \emph{Foundations and Trends{\textregistered} in Machine learning},
  3\penalty0 (1):\penalty0 1--122, 2011.

\bibitem[Breheny and Huang(2011)]{breheny2011coordinate}
Patrick Breheny and Jian Huang.
\newblock Coordinate descent algorithms for nonconvex penalized regression,
  with applications to biological feature selection.
\newblock \emph{The annals of applied statistics}, 5\penalty0 (1):\penalty0
  232, 2011.

\bibitem[B{\"u}hlmann and Van De~Geer(2011)]{buhlmann2011statistics}
Peter B{\"u}hlmann and Sara Van De~Geer.
\newblock \emph{Statistics for high-dimensional data: methods, theory and
  applications}.
\newblock Springer Science \& Business Media, 2011.

\bibitem[Cao et~al.(2019)Cao, Lin, and Li]{cao2018large}
Yuanpei Cao, Wei Lin, and Hongzhe Li.
\newblock Large covariance estimation for compositional data via
  composition-adjusted thresholding.
\newblock \emph{Journal of the American Statistical Association}, 114\penalty0
  (526):\penalty0 759--772, 2019.

\bibitem[Chen et~al.(2020)Chen, Liu, Mao, and Yang]{chen2020distributed}
Xi~Chen, Weidong Liu, Xiaojun Mao, and Zhuoyi Yang.
\newblock Distributed high-dimensional regression under a quantile loss
  function.
\newblock \emph{Journal of Machine Learning Research}, 21\penalty0
  (1):\penalty0 7432--7474, 2020.

\bibitem[Duan et~al.(2022)Duan, Ning, and Chen]{duan2022heterogeneity}
Rui Duan, Yang Ning, and Yong Chen.
\newblock Heterogeneity-aware and communication-efficient distributed
  statistical inference.
\newblock \emph{Biometrika}, 109\penalty0 (1):\penalty0 67--83, 2022.

\bibitem[Elgabli et~al.(2020{\natexlab{a}})Elgabli, Park, Bedi, Bennis, and
  Aggarwal]{elgabli2020gadmm}
Anis Elgabli, Jihong Park, Amrit~S Bedi, Mehdi Bennis, and Vaneet Aggarwal.
\newblock {GADMM}: Fast and communication efficient framework for distributed
  machine learning.
\newblock \emph{Journal of Machine Learning Research}, 21\penalty0
  (76):\penalty0 1--39, 2020{\natexlab{a}}.

\bibitem[Elgabli et~al.(2020{\natexlab{b}})Elgabli, Park, Bedi, Issaid, Bennis,
  and Aggarwal]{elgabli2020q}
Anis Elgabli, Jihong Park, Amrit~Singh Bedi, Chaouki~Ben Issaid, Mehdi Bennis,
  and Vaneet Aggarwal.
\newblock {Q-GADMM}: Quantized group admm for communication efficient
  decentralized machine learning.
\newblock \emph{IEEE Transactions on Communications}, 69\penalty0 (1):\penalty0
  164--181, 2020{\natexlab{b}}.

\bibitem[Fan and Li(2001)]{fan2001variable}
Jianqing Fan and Runze Li.
\newblock Variable selection via nonconcave penalized likelihood and its oracle
  properties.
\newblock \emph{Journal of the American statistical Association}, 96\penalty0
  (456):\penalty0 1348--1360, 2001.

\bibitem[Fan et~al.(2021)Fan, Guo, and Wang]{fan2021communication}
Jianqing Fan, Yongyi Guo, and Kaizheng Wang.
\newblock Communication-efficient accurate statistical estimation.
\newblock \emph{Journal of the American Statistical Association}, pages 1--11,
  2021.

\bibitem[Fan and Fan(2023)]{fan2021distributed}
Ye~Fan and Suning Fan.
\newblock Distributed adaptive lasso penalized generalized linear models for
  big data.
\newblock \emph{Communications in Statistics-Simulation and Computation},
  52\penalty0 (4):\penalty0 1679--1698, 2023.

\bibitem[Fan and Tang(2013)]{fan2013tuning}
Yingying Fan and Cheng~Yong Tang.
\newblock Tuning parameter selection in high dimensional penalized likelihood.
\newblock \emph{Journal of the Royal Statistical Society: Series B (Statistical
  Methodology)}, 75\penalty0 (3):\penalty0 531--552, 2013.

\bibitem[Friedman et~al.(2007)Friedman, Hastie, H{\"o}fling, and
  Tibshirani]{friedman2007pathwise}
Jerome Friedman, Trevor Hastie, Holger H{\"o}fling, and Robert Tibshirani.
\newblock Pathwise coordinate optimization.
\newblock \emph{The annals of applied statistics}, 1\penalty0 (2):\penalty0
  302--332, 2007.

\bibitem[Gao et~al.(2022)Gao, Liu, Wang, Wang, Yan, and Zhang]{gao2022review}
Yuan Gao, Weidong Liu, Hansheng Wang, Xiaozhou Wang, Yibo Yan, and Riquan
  Zhang.
\newblock A review of distributed statistical inference.
\newblock \emph{Statistical Theory and Related Fields}, 6\penalty0
  (2):\penalty0 89--99, 2022.

\bibitem[Giesen and Laue(2019)]{giesen2019combining}
Joachim Giesen and Soeren Laue.
\newblock Combining admm and the augmented lagrangian method for efficiently
  handling many constraints.
\newblock In \emph{Proceedings of the Twenty-Eighth International Joint
  Conference on Artificial Intelligence, {IJCAI-19}}, pages 4525--4531, 2019.

\bibitem[Giesen and Laue(2016)]{giesen2016distributed}
Joachim Giesen and S{\"o}ren Laue.
\newblock Distributed convex optimization with many convex constraints.
\newblock \emph{arXiv preprint arXiv:1610.02967}, 2016.

\bibitem[Gu and Zou(2020)]{gu2020sparse}
Yuwen Gu and Hui Zou.
\newblock Sparse composite quantile regression in ultrahigh dimensions with
  tuning parameter calibration.
\newblock \emph{IEEE Transactions on Information Theory}, 66\penalty0
  (11):\penalty0 7132--7154, 2020.

\bibitem[Gu et~al.(2018)Gu, Fan, Kong, Ma, and Zou]{gu2018admm}
Yuwen Gu, Jun Fan, Lingchen Kong, Shiqian Ma, and Hui Zou.
\newblock {ADMM} for high-dimensional sparse penalized quantile regression.
\newblock \emph{Technometrics}, 60\penalty0 (3):\penalty0 319--331, 2018.

\bibitem[Han et~al.(2022)Han, Huang, Lin, Liu, Qu, and Sun]{han2022robust}
Dongxiao Han, Jian Huang, Yuanyuan Lin, Lei Liu, Lianqiang Qu, and Liuquan Sun.
\newblock Robust signal recovery for high-dimensional linear log-contrast
  models with compositional covariates.
\newblock \emph{Journal of Business \& Economic Statistics}, pages 1--11, 2022.

\bibitem[Hu et~al.(2021)Hu, Jiao, Liu, Shi, and Wu]{hu2021distributed}
Aijun Hu, Yuling Jiao, Yanyan Liu, Yueyong Shi, and Yuanshan Wu.
\newblock Distributed quantile regression for massive heterogeneous data.
\newblock \emph{Neurocomputing}, 448:\penalty0 249--262, 2021.

\bibitem[Huang et~al.(2021)Huang, Lin, Zhang, and
  Zhang]{huang2021communication}
Zengfeng Huang, Xuemin Lin, Wenjie Zhang, and Ying Zhang.
\newblock Communication-efficient distributed covariance sketch, with
  application to distributed {PCA}.
\newblock \emph{Journal of Machine Learning Research}, 22\penalty0
  (1):\penalty0 3643--3680, 2021.

\bibitem[Issaid et~al.(2020)Issaid, Elgabli, Park, Bennis, and
  Debbah]{issaid2020communication}
Chaouki~Ben Issaid, Anis Elgabli, Jihong Park, Mehdi Bennis, and M{\'e}rouane
  Debbah.
\newblock Communication efficient distributed learning with censored,
  quantized, and generalized group {ADMM}.
\newblock \emph{arXiv preprint arXiv:2009.06459}, 2020.

\bibitem[Jordan et~al.(2019)Jordan, Lee, and Yang]{jordan2018communication}
Michael~I Jordan, Jason~D Lee, and Yun Yang.
\newblock Communication-efficient distributed statistical inference.
\newblock \emph{Journal of the American Statistical Association}, 114\penalty0
  (526):\penalty0 668--681, 2019.

\bibitem[Kairouz et~al.(2021)Kairouz, McMahan, Avent, Bellet, Bennis, Bhagoji,
  Bonawitz, Charles, Cormode, Cummings, et~al.]{kairouz2021advances}
Peter Kairouz, H~Brendan McMahan, Brendan Avent, Aur{\'e}lien Bellet, Mehdi
  Bennis, Arjun~Nitin Bhagoji, Kallista Bonawitz, Zachary Charles, Graham
  Cormode, Rachel Cummings, et~al.
\newblock Advances and open problems in federated learning.
\newblock \emph{Foundations and Trends{\textregistered} in Machine Learning},
  14\penalty0 (1--2):\penalty0 1--210, 2021.

\bibitem[Kone{\v{c}}n{\`y} et~al.(2016)Kone{\v{c}}n{\`y}, McMahan, Yu,
  Richt{\'a}rik, Suresh, and Bacon]{konevcny2016federated}
Jakub Kone{\v{c}}n{\`y}, H~Brendan McMahan, Felix~X Yu, Peter Richt{\'a}rik,
  Ananda~Theertha Suresh, and Dave Bacon.
\newblock Federated learning: Strategies for improving communication
  efficiency.
\newblock \emph{arXiv preprint arXiv:1610.05492}, 2016.

\bibitem[Li et~al.(2022)Li, Srinivasan, Chen, and Xue]{li2022robust}
Danning Li, Arun Srinivasan, Qian Chen, and Lingzhou Xue.
\newblock Robust covariance matrix estimation for high-dimensional
  compositional data with application to sales data analysis.
\newblock \emph{Journal of Business \& Economic Statistics}, pages 1--11, 2022.

\bibitem[Li and Zhao(2022)]{li2021communication}
Mengyu Li and Junlong Zhao.
\newblock Communication-efficient distributed linear discriminant analysis for
  binary classification.
\newblock \emph{Statistica Sinica}, 32:\penalty0 1343--1361, 2022.

\bibitem[Lin et~al.(2014)Lin, Shi, Feng, and Li]{lin2014variable}
Wei Lin, Pixu Shi, Rui Feng, and Hongzhe Li.
\newblock Variable selection in regression with compositional covariates.
\newblock \emph{Biometrika}, 101\penalty0 (4):\penalty0 785--797, 2014.

\bibitem[Liu et~al.(2023)Liu, Zhao, and Pan]{liu2022communication}
Zhan Liu, Xiaoluo Zhao, and Yingli Pan.
\newblock Communication-efficient distributed estimation for high-dimensional
  large-scale linear regression.
\newblock \emph{Metrika}, 86\penalty0 (4):\penalty0 455--485, 2023.

\bibitem[Lu et~al.(2019)Lu, Shi, and Li]{lu2019generalized}
Jiarui Lu, Pixu Shi, and Hongzhe Li.
\newblock Generalized linear models with linear constraints for microbiome
  compositional data.
\newblock \emph{Biometrics}, 75\penalty0 (1):\penalty0 235--244, 2019.

\bibitem[Ma et~al.(2022)Ma, Wang, and Zhou]{ma2022statistical}
Xuejun Ma, Shaochen Wang, and Wang Zhou.
\newblock Statistical inference in massive datasets by empirical likelihood.
\newblock \emph{Computational Statistics}, 37\penalty0 (3):\penalty0
  1143--1164, 2022.

\bibitem[Mateos et~al.(2010)Mateos, Bazerque, and
  Giannakis]{mateos2010distributed}
Gonzalo Mateos, Juan~Andr{\'e}s Bazerque, and Georgios~B Giannakis.
\newblock Distributed sparse linear regression.
\newblock \emph{IEEE Transactions on Signal Processing}, 58\penalty0
  (10):\penalty0 5262--5276, 2010.

\bibitem[Mishra and M{\"u}ller(2022)]{mishra2022robust}
Aditya Mishra and Christian~L M{\"u}ller.
\newblock Robust regression with compositional covariates.
\newblock \emph{Computational Statistics \& Data Analysis}, 165:\penalty0
  107315, 2022.

\bibitem[Nedi{\'c} et~al.(2018)Nedi{\'c}, Olshevsky, and
  Rabbat]{nedic2018network}
Angelia Nedi{\'c}, Alex Olshevsky, and Michael~G Rabbat.
\newblock Network topology and communication-computation tradeoffs in
  decentralized optimization.
\newblock \emph{Proceedings of the IEEE}, 106\penalty0 (5):\penalty0 953--976,
  2018.

\bibitem[Pan(2021)]{pan2021distributed}
Yingli Pan.
\newblock Distributed optimization and statistical learning for large-scale
  penalized expectile regression.
\newblock \emph{Journal of the Korean Statistical Society}, 50\penalty0
  (1):\penalty0 290--314, 2021.

\bibitem[Pan et~al.(2022)Pan, Xu, Wei, Wang, and Liu]{pan2022efficient}
Yingli Pan, Kaidong Xu, Sha Wei, Xiaojuan Wang, and Zhan Liu.
\newblock Efficient distributed optimization for large-scale high-dimensional
  sparse penalized huber regression.
\newblock \emph{Communications in Statistics-Simulation and Computation}, pages
  1--20, 2022.

\bibitem[Shi et~al.(2021)Shi, Qin, Zhu, and Zhu]{shi2021communication}
Jianwei Shi, Guoyou Qin, Huichen Zhu, and Zhongyi Zhu.
\newblock Communication-efficient distributed m-estimation with missing data.
\newblock \emph{Computational Statistics \& Data Analysis}, 161:\penalty0
  107251, 2021.

\bibitem[Shi et~al.(2016)Shi, Zhang, and Li]{shi2016regression}
Pixu Shi, Anru Zhang, and Hongzhe Li.
\newblock Regression analysis for microbiome compositional data.
\newblock \emph{The Annals of Applied Statistics}, 10\penalty0 (2):\penalty0
  1019--1040, 2016.

\bibitem[Sohrab(2003)]{sohrab2003basic}
Houshang~H Sohrab.
\newblock \emph{Basic real analysis}, volume 231.
\newblock Springer, 2003.

\bibitem[Tan et~al.(2022)Tan, Battey, and Zhou]{tan2022communication}
Kean~Ming Tan, Heather Battey, and Wen-Xin Zhou.
\newblock Communication-constrained distributed quantile regression with
  optimal statistical guarantees.
\newblock \emph{Journal of Machine Learning Research}, 23:\penalty0 1--61,
  2022.

\bibitem[Tibshirani(1996)]{tibshirani1996regression}
Robert Tibshirani.
\newblock Regression shrinkage and selection via the lasso.
\newblock \emph{Journal of the Royal Statistical Society: Series B (Satistical
  Methodology)}, 58\penalty0 (1):\penalty0 267--288, 1996.

\bibitem[Volgushev et~al.(2019)Volgushev, Chao, and
  Cheng]{volgushev2019distributed}
Stanislav Volgushev, Shih-Kang Chao, and Guang Cheng.
\newblock Distributed inference for quantile regression processes.
\newblock \emph{The Annals of Statistics}, 47\penalty0 (3):\penalty0
  1634--1662, 2019.

\bibitem[Wang et~al.(2021)Wang, Li, and Zhang]{wang2021robust}
Kangning Wang, Shaomin Li, and Benle Zhang.
\newblock Robust communication-efficient distributed composite quantile
  regression and variable selection for massive data.
\newblock \emph{Computational Statistics \& Data Analysis}, 161:\penalty0
  107262, 2021.

\bibitem[Wang et~al.(2022{\natexlab{a}})Wang, Zhang, Alenezi, and
  Li]{wang2022communication}
Kangning Wang, Benle Zhang, Fayadh Alenezi, and Shaomin Li.
\newblock Communication-efficient surrogate quantile regression for
  non-randomly distributed system.
\newblock \emph{Information Sciences}, 588:\penalty0 425--441,
  2022{\natexlab{a}}.

\bibitem[Wang et~al.(2022{\natexlab{b}})Wang, Zhang, Sun, and
  Li]{wang2022efficient}
Kangning Wang, Benle Zhang, Xiaofei Sun, and Shaomin Li.
\newblock Efficient statistical estimation for a non-randomly distributed
  system with application to large-scale data neural network.
\newblock \emph{Expert Systems with Applications}, 197:\penalty0 116698,
  2022{\natexlab{b}}.

\bibitem[Wang and Zhao(2017)]{wang2017structured}
Tao Wang and Hongyu Zhao.
\newblock Structured subcomposition selection in regression and its application
  to microbiome data analysis.
\newblock \emph{The Annals of Applied Statistics}, 11\penalty0 (2):\penalty0
  771--791, 2017.

\bibitem[Wright(2015)]{wright2015coordinate}
Stephen~J Wright.
\newblock Coordinate descent algorithms.
\newblock \emph{Mathematical programming}, 151\penalty0 (1):\penalty0 3--34,
  2015.

\bibitem[Wu et~al.(2022)Wu, Huang, and Wang]{wu2022network}
Shuyuan Wu, Danyang Huang, and Hansheng Wang.
\newblock Network gradient descent algorithm for decentralized federated
  learning.
\newblock \emph{Journal of Business \& Economic Statistics}, \penalty0
  (just-accepted):\penalty0 1--31, 2022.

\bibitem[Yang and Wang(2023)]{yang2022communication}
Yaohong Yang and Lei Wang.
\newblock Communication-efficient sparse composite quantile regression for
  distributed data.
\newblock \emph{Metrika}, 86\penalty0 (3):\penalty0 261--283, 2023.

\bibitem[Yu et~al.(2017)Yu, Lin, and Wang]{yu2017parallel}
Liqun Yu, Nan Lin, and Lan Wang.
\newblock A parallel algorithm for large-scale nonconvex penalized quantile
  regression.
\newblock \emph{Journal of Computational and Graphical Statistics}, 26\penalty0
  (4):\penalty0 935--939, 2017.

\bibitem[Yu et~al.(2022)Yu, Chao, and Cheng]{yu2022distributed}
Yang Yu, Shih-Kang Chao, and Guang Cheng.
\newblock Distributed bootstrap for simultaneous inference under high
  dimensionality.
\newblock \emph{Journal of Machine Learning Research}, 23\penalty0
  (195):\penalty0 1--77, 2022.

\bibitem[Zhang et~al.(2013)Zhang, Duchi, and
  Wainwright]{zhang2013communication}
Yuchen Zhang, John~C Duchi, and Martin~J Wainwright.
\newblock Communication-efficient algorithms for statistical optimization.
\newblock \emph{Journal of Machine Learning Research}, 14:\penalty0 3321--3363,
  2013.

\bibitem[Zhou et~al.(2021)Zhou, Yu, Ma, Tian, and Fan]{zhou2021communication}
Ping Zhou, Zhen Yu, Jingyi Ma, Maozai Tian, and Ye~Fan.
\newblock Communication-efficient distributed estimator for generalized linear
  models with a diverging number of covariates.
\newblock \emph{Computational Statistics \& Data Analysis}, 157:\penalty0
  107154, 2021.

\bibitem[Zhu et~al.(2021)Zhu, Li, and Wang]{zhu2021least}
Xuening Zhu, Feng Li, and Hansheng Wang.
\newblock Least-square approximation for a distributed system.
\newblock \emph{Journal of Computational and Graphical Statistics}, 30\penalty0
  (4):\penalty0 1004--1018, 2021.

\bibitem[Zou(2006)]{zou2006adaptive}
Hui Zou.
\newblock The adaptive lasso and its oracle properties.
\newblock \emph{Journal of the American statistical association}, 101\penalty0
  (476):\penalty0 1418--1429, 2006.

\bibitem[Zou and Li(2008)]{zou2008one}
Hui Zou and Runze Li.
\newblock One-step sparse estimates in nonconcave penalized likelihood models.
\newblock \emph{The Annals of statistics}, 36\penalty0 (4):\penalty0
  1509--1533, 2008.

\end{thebibliography}

\end{document}